\documentclass[lettersize,journal]{IEEEtran}
\usepackage{amsmath,amsfonts}
\usepackage{algorithmic}
\usepackage{algorithm}
\usepackage{array}
\usepackage[caption=false,font=normalsize,labelfont=sf,textfont=sf]{subfig}
\usepackage{textcomp}
\usepackage{stfloats}
\usepackage{url}
\usepackage{verbatim}
\usepackage{graphicx}
\usepackage{cite}
\usepackage{hyperref}
\usepackage{booktabs}

\begin{document}

\title{Automating Infrastructure Surveying: A Framework for Geometric Measurements and Compliance Assessment Using Point Cloud Data}

\author{
Amin Ghafourian, Andrew Lee, Dechen Gao, Tyler Beer, Kin Yen, Iman Soltani,~\IEEEmembership{Member,~IEEE} \\\vspace{6pt}LARA Research, Department of Mechanical and Aerospace Engineering \\ University of California, Davis
        %<-this % stops a space
\thanks{This work was supported by the California Department of Transportation (Caltrans). Corresponding Author: Iman soltani (isoltani@ucdavis.edu).}% <-this % stops a space
% \thanks{Manuscript received ..., 2025; revised ..., 2025.}}
}

% \markboth{IEEE Transactions on Automation Science and Engineering,~Vol.~??, No.~??, August~2025}%


\maketitle
\begin{abstract}
Automation can play a prominent role in improving efficiency, accuracy, and scalability in infrastructure surveying and assessing construction and compliance standards. This paper presents a framework for automation of geometric measurements and compliance assessment using point cloud data. The proposed approach integrates deep learning-based detection and segmentation, in conjunction with geometric and signal processing techniques, to automate surveying tasks. As a proof of concept, we apply this framework to automatically evaluate the compliance of curb ramps with the Americans with Disabilities Act (ADA), demonstrating the utility of point cloud data in survey automation. The method leverages a newly collected, large annotated dataset of curb ramps, made publicly available as part of this work, to facilitate robust model training and evaluation. Experimental results, including comparison with manual field measurements of several ramps, validate the accuracy and reliability of the proposed method, highlighting its potential to significantly reduce manual effort and improve consistency in infrastructure assessment. Beyond ADA compliance, the proposed framework lays the groundwork for broader applications in infrastructure surveying and automated construction evaluation, promoting wider adoption of point cloud data in these domains. The annotated database, manual ramp survey data, and developed algorithms are publicly available on the project’s GitHub page: \href{https://github.com/Soltanilara/SurveyAutomation}{github.com/Soltanilara/SurveyAutomation}.  
\end{abstract}

\begin{ntp}
    Routine checks of infrastructure, such as road and pavement assets, still require field crews to perform manual field inspection and measurements, even though many agencies already own substantial digital twins, like dense mobile LiDAR surveys. This paper describes a multi-stage technique that leverages point clouds to reliably automate the inspection and measurement tasks, significantly relaxing the need for labor-intensive, time-consuming manual procedures. The method combines advanced computer vision techniques for asset detection and visual component segmentation with mathematical and geometric modeling and rule-based pipelines, as well as classical machine learning techniques, to rigorously incorporate priors and constraints associated with the survey task and asset. The proposed technique is applied to the task of surveying ADA curb ramps, showing a substantial reduction in need for manual field measurements and the possibility of wide adoption. We are also publicly releasing the source codes and datasets for future developments by other researchers and practitioners. The method is mainly limited by the quality of the digital data, as well as the availability of high-quality human annotations used for training computer vision models.
\end{ntp}

\vspace{28pt}
\section{Introduction}
\IEEEPARstart{S}{urveying} plays a pivotal role in city management and the maintenance of urban infrastructure, ensuring the functionality, safety, and accessibility of public assets. Accurate and timely surveys are essential for assessing the condition of infrastructure, identifying signs of deterioration, and planning necessary repairs or upgrades~\cite{biondini2023life}. These activities also serve a critical function in verifying compliance with design specifications, construction standards, and regulatory frameworks~\cite{biondini2023life}. For instance, adherence to the ADA requirements ensures equitable access to public infrastructure for all individuals. By systematically documenting infrastructure conditions and compliance, surveying not only supports effective resource allocation but also minimizes potential liabilities associated with non-compliance or unsafe conditions. As cities expand and age, the demand for efficient and accurate surveying continues to grow~\cite{doi:10.1061/(ASCE)0887-3801(2008)22:3(216)}.

Surveying is an exceptionally labor-intensive process due to the sheer scale and diversity of infrastructure and building assets that require regular assessment~\cite{biondini2023life,Economicefficiency}. Urban environments encompass a vast array of assets, including roadways, sidewalks, bridges, parking facilities, drainage systems, and utilities, each with its own unique set of compliance requirements and measurement criteria. For example, ensuring ADA compliance at pedestrian crossings involves assessing thousands of curb ramps, each requiring measurements of multiple dimensions and slopes~\cite{jacobs2009ada}. Similarly, evaluating the structural integrity of bridges demands detailed inspections of girders, bearings, and joints~\cite{fhwa2022nbi}, while roadway surveys may require precise measurement of pavement conditions, lane markings, and traffic signage~\cite{pierce2013practical,mallela2013mire}. Given the broad variety and geographic spread of these assets, their survey demands considerable manpower and time, resulting in high costs and delays. As such, the burden of manual surveying is becoming increasingly unsustainable~\cite{torres2018automation}, emphasizing the urgent need for automation.

In recent years, cities have increasingly turned to mobile terrestrial scanning technologies to document and manage urban infrastructure more effectively. These systems generate high-resolution, georeferenced point cloud data paired with registered images, providing a detailed and comprehensive digital representation of cityscapes, commonly referred to as digital twins. These digital twins serve as invaluable resources for infrastructure management, offering unprecedented opportunities for monitoring, planning, and decision-making. However, despite the growing availability of such rich datasets, their potential remains largely untapped for automating critical surveying tasks. Current efforts have predominantly focused on visualization and manual data interpretation, leaving a significant gap in leveraging these resources to streamline and enhance infrastructure assessment processes. By automating the analysis and extraction of actionable insights, digital twins can transform urban surveying, delivering efficiencies previously out of reach.

Recent advancements in artificial intelligence (AI), particularly in machine learning (ML) and deep learning (DL), present transformative opportunities for automating surveying tasks, offering the potential to save millions of dollars annually. These technologies excel in tasks such as feature/object detection, segmentation, and classification, making them valuable tools for processing large-scale data efficiently. However, sole reliance on end-to-end AI-based solutions is not sufficient for the quantitative demands of surveying, where precision and accuracy are paramount. Such end-to-end methods often require vast amounts of labeled training data, which are time-consuming to generate and may still fall short of achieving the stringent accuracy needed for many surveying applications. 

We propose a hybrid framework that combines the strengths of modern AI tools with conventional geometric analysis techniques to automate infrastructure surveying. Our approach leverages AI for robust asset detection and segmentation, while relying on classical geometric methods for precise quantitative measurements, thus reducing dependence on the precision of AI outputs. By integrating the scalability of AI with the accuracy and reliability of geometric analysis, the framework offers a generalizable solution for diverse surveying applications.

\begin{figure}[!t]
\centering
\includegraphics[width=\linewidth]{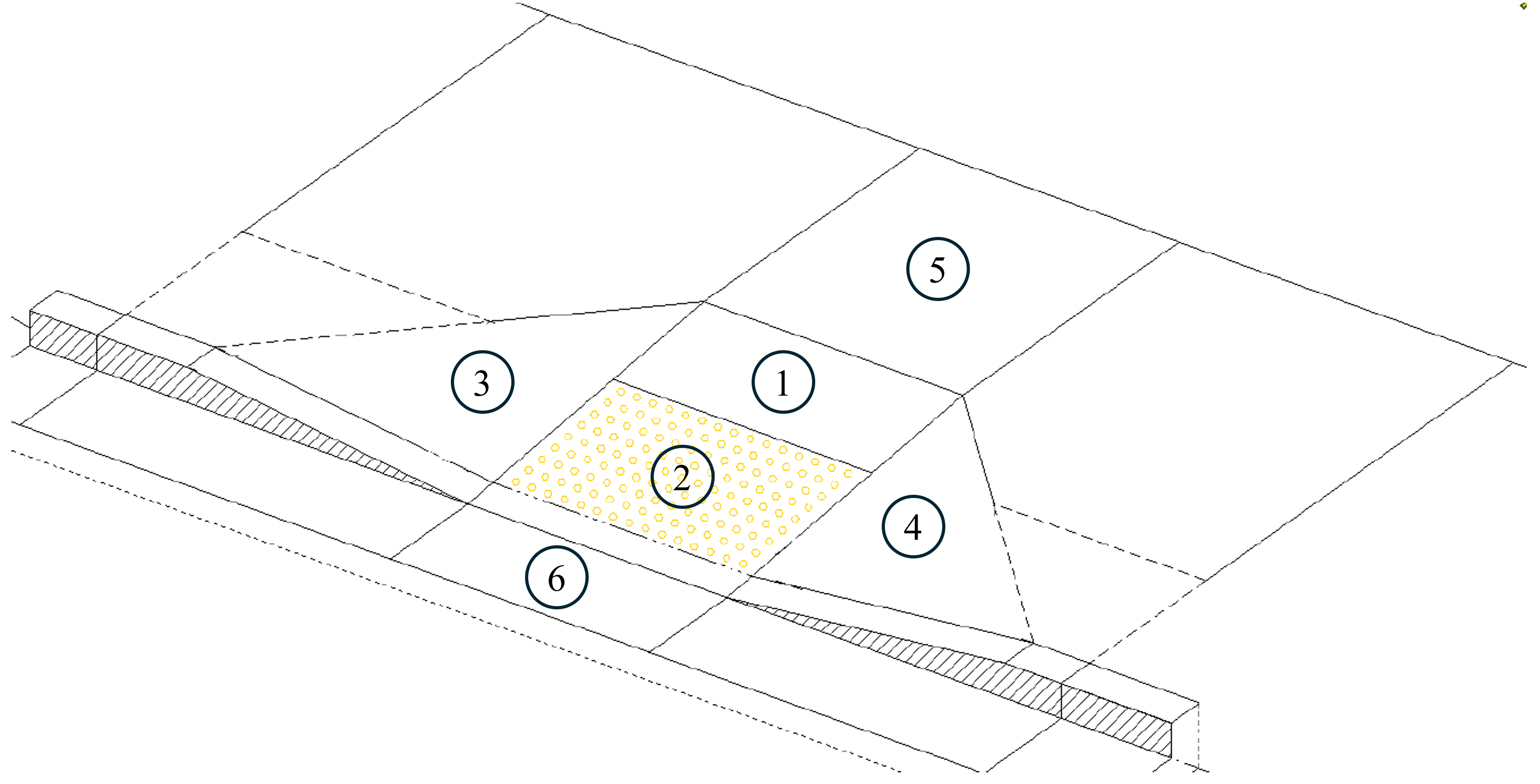}
\caption{A prototypical ramp representing ramp types A, D, and E~\cite{CEM5773ADE}. Center ramp (1), warning surface (2), left flare (3), right flare (4), landing (5), and gutter (6) are denoted on the figure. Assessing ADA compliance requires multiple measurements across each ramp component, making the manual process labor-intensive, time-consuming, and costly.}
\label{fig: ramp_components}
\end{figure}

We demonstrate the effectiveness of our approach through the use case of assessing curb ramps at pedestrian crossings, an application that encapsulates the core challenges of infrastructure surveying. ADA compliance assessment for ramps requires numerous precise geometric measurements, including multiple slopes, cross-slopes, widths, and surface deviations across several ramp components such as the center ramp, warning surface, flares, gutter, and landing areas (Fig.~\ref{fig: ramp_components}). The required level of detail and accuracy makes manual assessment of ramps labor-intensive and time-consuming, creating a strong case for automation. Moreover, the complexity and variability of curb ramp designs make this task a rich and meaningful testbed for evaluating the generalizability and robustness of the proposed framework.

To support this effort, we introduce a dataset of over 1600 annotated and segmented ramps, one of the first of its kind, which we are making publicly available. The data includes both field measurements and expert-validated digital point cloud annotations, providing a strong benchmark for evaluation. By sharing datasets, ground-truth measurements, and open-source algorithms, this work aims to accelerate further research and foster reproducibility in automated infrastructure surveying.

The remainder of this paper is organized as follows: Section~\ref{sec: priorWork} discusses related works in the space of infrastructure inspection and survey automation. In Section~\ref{sec: hybridFramework} we introduce the proposed hybrid framework for survey automation, and in Section~\ref{sec: caseStudy} we demonstrate the application of the proposed framework to ADA ramp assets. Section~\ref{sec: experiments} details the experiments and results of the case study. The outcomes, caveats, and potential improvements are discussed in Section~\ref{sec: discussion}. Section~\ref{sec: conclusion} concludes the paper.

\section{Prior Work}\label{sec: priorWork}
While there has been limited direct research on automating surveying tasks, several related studies have explored the use of digital data, such as point clouds, in ways that lay important groundwork for such automation. A notable study \cite{10641055} introduced a deep learning-based approach for point cloud classification using transformers for applications such as urban planning and infrastructure management. In the realm of bridge inspection, \cite{rs12223757} compared three deep learning models, PointNet~\cite{8099499}, PointCNN~\cite{3326943.3327020}, and Dynamic Graph Convolutional Neural Network (DGCNN)~\cite{10.1145/3326362}, for classifying bridge components from point cloud data, highlighting the potential of deep learning in automating structural inspections. Additionally, \cite{Kumar17072020} proposed a multi-faceted multi-object convolutional neural network (MMCN) combined with a support vector machine (SVM)~\cite{suthaharan2016support} for the fully automated classification of objects such as houses, trees, poles and cars in highly dense Mobile Terrestrial Laser Scanning (MTLS) 3D point cloud data. \cite{10887345} proposes an automatic, coarse-to-fine point cloud clustering method for 3D surface defect diagnosis on flat steel surfaces. \cite{10988584} presents a specialized attention-based recognition network for building information modeling (BIM) of existing pipe systems, addressing the challenges of complex geometry and mobile deployment by using a flexible decoder design and a geometry-aware convolution pyramid. In another effort, \cite{9259076} developed a hybrid deep learning framework combining CNN and LSTM architectures with time–frequency signal processing techniques to enable accurate and resource-efficient damage detection for real-time structural health monitoring of large-scale infrastructure such as bridges \cite{9374750}. 

A recent study investigated the efficacy of various point-cloud-based deep learning models, including PointNet, PointNet++~\cite{qi2017pointnetdeephierarchicalfeature}, 3D-CNNs, and PointCNN, in automating roadway health assessments, utilizing data from multiple roadways in Southern California~\cite{Zhang2023LiDAR}. Furthermore, transformer-based approaches have been applied to automate the extraction of rural multilane highway infrastructure elements from LiDAR data, achieving high accuracy in classifying various highway components~\cite{doi:10.1139/cjce-2024-0312}. Moreover, the U.S. Geological Survey (USGS) has initiated studies utilizing deep learning technologies to automate the classification of point clouds from the 3D Elevation Program (3DEP) LiDAR data, aiming to refine and enrich existing datasets~\cite{10641055}.

Other research has employed UAV-based imaging and photogrammetry to inspect and model roadways, bridges, tunnels, and even powerlines, thereby complementing LiDAR-based approaches in infrastructure surveying. For instance, \cite{https://doi.org/10.1111/j.1467-8667.2011.00727.x} deployed a UAV-based imaging system to generate high-resolution 3D reconstructions of unpaved roads, enabling more precise measurement of surface distresses. In a similar vein, \cite{SIEBERT20141} demonstrated how UAV photogrammetry can be used for surveying and monitoring earthwork projects, showing that digital datasets can expedite survey workflows while reducing the need for manual ground measurements. \cite{RAKHA2018252} conducted a comprehensive review of the applications of unmanned aerial systems for built-environment inspection tasks, highlighting how drone-based data acquisition can streamline site assessments and support proactive infrastructure maintenance efforts. In addition to data processing, this line of work was extended to the development of bottom-up flight planning frameworks for large-scale infrastructure inspection, which optimizes performance objectives across multiple time scales and overcomes the inefficiencies of traditional top-down mission design in applications such as powerline, road, and railway inspection \cite{9374750}.

Robotic systems equipped with cameras and scanning devices have been investigated for structural monitoring in high-risk environments, such as deteriorating bridges and tunnels \cite{6917066, 6705706, 10855578}. \cite{doi:10.1061/(ASCE)IS.1943-555X.0000353} reviewed various robotic inspection platforms that collect image and laser-scanned data to detect cracks, corrosion, or other anomalies in aging infrastructure. Such approaches have also been extended to marine environments, where uncrewed surface vessels (USVs) are deployed to collect bathymetry data and assist in surveying underwater infrastructure, including bridges \cite{kumar2025designimplementationdualuncrewed}.

These examples underscore how the expanding use of digital data, from LiDAR scans to aerial imagery and underwater bathymetry, is steadily transforming traditional surveying and inspection practices. This trend is paving the way for fully automated surveying workflows, presenting significant opportunities for research and real-world deployment.

\section{A Data-Driven Hybrid Framework for Automated Surveying Using Machine Learning and Geometric Analysis}\label{sec: hybridFramework}
\begin{figure}[!t]
\centering
\includegraphics[width=\linewidth]{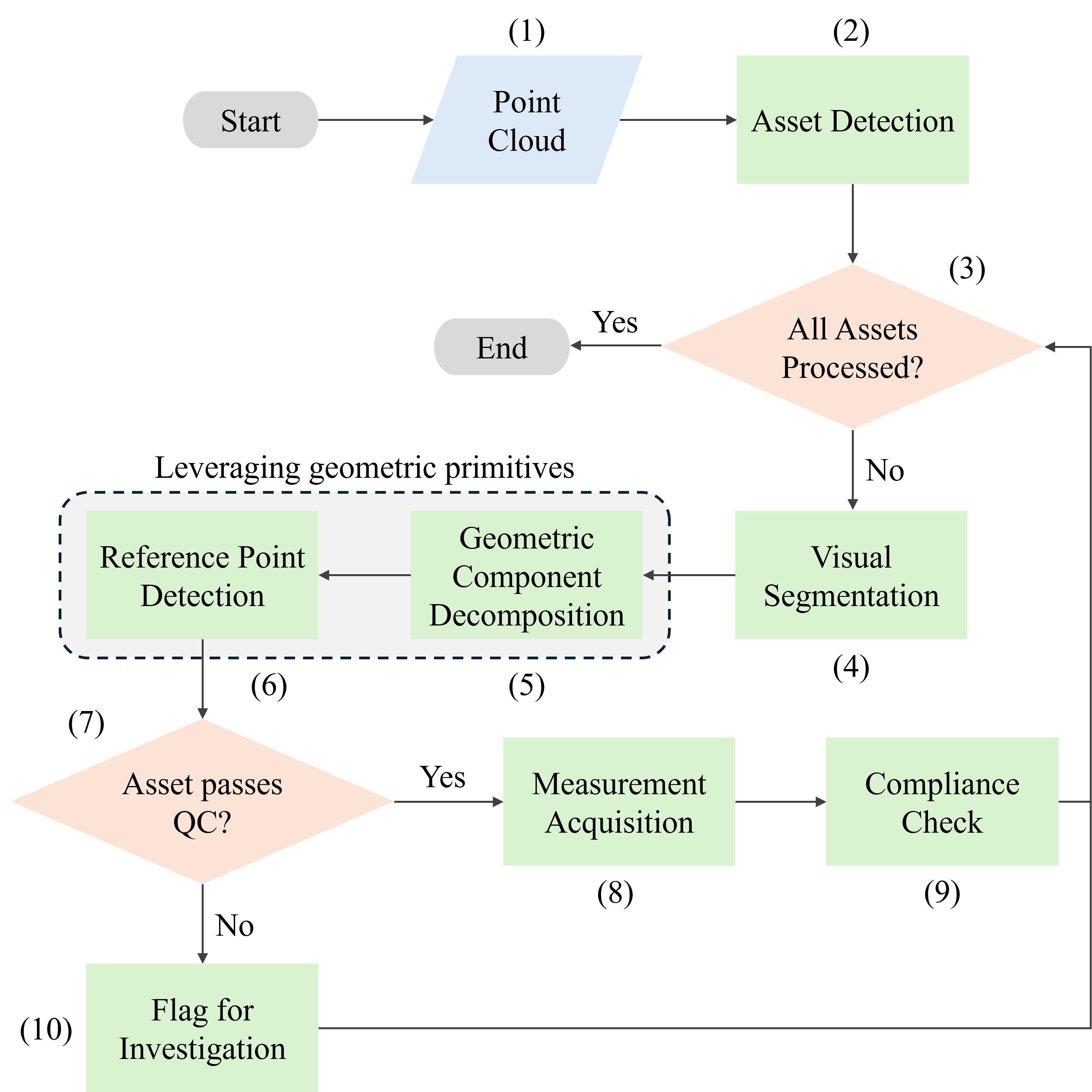}
\caption{Flowchart of the proposed infrastructure asset compliance assessment automation method based on point cloud data.}
\label{fig: flowchart}
\end{figure}

We present a structured methodology that integrates modern machine learning techniques with classical geometric analysis to enable scalable and precise automation of surveying tasks in infrastructure management. This framework begins with identifying assets of interest in MTLS data, leveraging advanced object detection techniques. Once the asset is located, segmentation methods are applied to isolate its specific components of interest. Following segmentation, analytical techniques are employed to decompose the asset into its fundamental geometric primitives, such as planes, lines, and points. These primitives enable the determination of critical geometric references, including intersections and boundaries, which serve as anchors for making quantitative measurements. This top-down approach integrates the scalability of machine learning–based data processing with the accuracy of geometric methods to enable automation of surveying tasks, overcoming the limitations inherent in using either approach alone. A flowchart of the proposed automation framework is shown in Fig.~\ref{fig: flowchart}.

\subsection{Asset Detection}
Following point cloud data collection, the next step in the proposed framework is the identification of assets of interest within the MTLS data. This process involves applying ML-based object detection techniques, leveraging either the 2D image data, 2D projections of 3D point cloud, raw 3D point clouds, or a combination. For 2D approaches, state-of-the-art object detection models, such as DeTR  \cite{carion2020end} and Faster R-CNN \cite{ren2016faster}, have demonstrated robust performance in identifying objects from images. These methods can also be applied to 2D projections of point clouds \cite{liu2021bev, li2024bevnext, peng2023bevsegformer}, such as bird’s-eye views or height maps, offering a simple yet effective way to leverage 2D object detection tools for 3D data. For 3D object detection, models like PointRCNN \cite{shi2019pointrcnn} and VoteNet \cite{qi2019deep} have emerged as leading solutions, specifically designed to operate directly on point cloud data. Hybrid approaches that combine image and point cloud data, such as Multi-View CNNs or multimodal fusion networks \cite{9127813}, further enhance detection accuracy by integrating complementary features from both data modalities. Accurate asset detection is essential, as it underpins all downstream steps by ensuring that only the relevant portions of the dataset are processed for measurement extraction.

\subsection{Visual Segmentation}
Once the asset of interest is identified, the next step involves segmenting it into its visually distinct constituent parts to simplify further analysis and measurement. Similar to the detection stage, segmentation can be approached using either 2D or 3D techniques, depending on the data modality and application requirements. For 2D data, semantic segmentation models like DeepLabV3+ \cite{9186684}, U-Net \cite{WANG2021106373,10.1007/978-3-319-24574-4_28}, and more recently SAM (segment anything) \cite{Kirillov_2023_ICCV} have proven highly effective, offering pixel-level classification in registered images or 2D projections of point clouds. On the other hand, 3D segmentation techniques, such as those based on PointNet++ \cite{qi2017pointnetdeephierarchicalfeature}, KPConv (Kernel Point Convolution) \cite{thomas2019kpconvflexibledeformableconvolution}, or RandLA-Net \cite{hu2020randlanetefficientsemanticsegmentation}, directly operate on raw point cloud data. These methods excel in parsing complex, unstructured datasets to isolate meaningful regions. Recent advancements also highlight the use of hybrid segmentation approaches, where 2D projections are segmented first, and the results are mapped back to the 3D domain to enhance computational efficiency and accuracy \cite{lyu2020learningsegment3dpoint, yang20242d3dinterlacedtransformerpoint,guo2024samguidedgraphcut3d}. 

\subsection{Geometric Component Decomposition}
The next step in the proposed framework focuses on extracting geometric primitives, leveraging the inherent assumption that man-made structures are often composed of simple, well-defined geometric elements such as planes, edges, and lines. The segmentation results from the previous stage serve as input, providing groups of points that can be fitted with these geometric primitives. Simple techniques, such as plane fitting, edge detection, and geometric modeling, are employed to identify the most reliable primitives while filtering out noise and outliers in the data. By grounding the process in the known geometric structure of the asset, the extracted primitives offer a more accurate representation than segmentation results alone, which rely solely on visual cues. This ensures that the asset’s building blocks are precisely defined. This approach is particularly critical for addressing the imperfections and noise inherent in MTLS data or segmentation outputs. Untreated noise can significantly impact the accuracy of quantitative measurements, but extracting primitives based on a geometric model mitigates these errors, enabling a cleaner and more structured representation of the assets of interest. This stage lays a foundation for the identification of geometric references in the next step.

\subsection{Geometric Reference Detection}
The next step in the proposed framework involves extracting geometric references, enabling the calculation of key quantitative measurements such as slopes, widths, surface deviations, and other essential parameters. Building on the geometric primitives extracted in the previous step, plane intersections are used to derive lines, and points are identified by further refinement—such as through line intersections or by optimizing geometric criteria like distances from lines or planes. These geometric references provide a minimalistic and abstract representation of the asset, suitable for precise measurement extraction.

\subsection{Quality Control}
To ensure the reliability of extracted measurements in any automated surveying task, it is essential to incorporate quality control (QC) measures at multiple stages of the pipeline. These QC steps play a critical role in identifying potential sources of error and flagging outputs that may warrant manual inspection.

A fundamental challenge in automated surveying is that the reliability of extracted measurements cannot be definitively verified without on-site validation. However, systematic QC procedures can help ensure that the data and intermediate results are within expected ranges, thereby increasing confidence in the final outputs. Assuming the availability of a model of the asset, certain geometric bounds can be defined. As a few instances, allowable surface deviations, expected angle ranges between planes or lines, and distances between geometric references can be used to identify anomalies. Such anomalies may indicate severely distorted assets requiring urgent attention or errors in the processing pipeline that compromise measurement accuracy.

In our proposed framework, QC measures are integrated at two key stages: once following the extraction of geometric references, and again at the final stage when measurements are computed. These checkpoints help catch both early and late-stage issues and can be customized or extended based on the specific needs of a given surveying application. 

\section{A Case Study: Automated ADA Ramp Compliance Assessment}\label{sec: caseStudy}
In recent years, California has faced significant legal challenges regarding the non-compliance of its infrastructure with the ADA. In 2009, the California Department of Transportation (Caltrans) settled a lawsuit by pledging \$1.1 billion to improve sidewalk access statewide, addressing deficiencies and non-compliances \cite{CDR_v_Caltrans}. Similarly, in 2015, the City of Los Angeles resolved a landmark class-action lawsuit by committing \$1.4 billion over 30 years to repair sidewalks and install curb ramps that ensure compliance with ADA requirements \cite{Willits_v_LA}. These settlements represent some of the largest disability access-related agreements in U.S. history, highlighting the critical need for rigorous compliance with ADA standards to ensure accessibility and safety for all users.

As a case study, here we demonstrate and evaluate the proposed framework by applying it to the task of compliance assessment for ADA curb ramps at pedestrian crossings. This survey task provides an ideal example, given its reliance on precise geometric measurements and its labor-intensive nature, which highlights an urgent need for automation.

\subsection{Manual Procedure for ADA Compliance Assessment}
Conventionally, ramp measurements are conducted manually in the field by trained personnel. Depending on the ramp type, specific parameters such as slopes, cross slopes, widths, surface deviations, and dimensions of features like warning surfaces, flares, gutter and landing areas (see Fig.~\ref{fig: ramp_components}) must be recorded and compared against approved reference values~\cite{CEM5773ADE}. For this study, we focus on three primary ramp types, referred to as types A, D, and E, requiring a unique set of measurements based on their geometric characteristics. For each ramp, field personnel are required to take dozens of distinct measurements. These measurements are typically performed using tools such as inclinometers and measuring tapes, often relying on visual cues to identify boundaries. Once the measurements are collected, the data is populated into a standardized PDF form~\cite{CEM5773ADE}, which serves as an official record for compliance assessment. While thorough, this method is highly labor-intensive, error-prone, and constrained by the subjective interpretation of field personnel—especially when clear boundaries are absent or not accurately represented by the visual component separations that guide the operator’s judgment.

\subsection{Dataset}
In recent years, departments of transportation (DOTs) across the United States, including Caltrans, have increasingly adopted advanced technologies such as MTLS and STLS (Static Terrestrial Laser Scanning) to create detailed digital representations of urban infrastructure. This shift towards digitalization has been driven by the need for accurate, efficient, and safe methods of managing transportation assets. Caltrans, for instance, owns and operates several MTLS systems such as the Trimble MX9 and Trimble MX50. These systems have been instrumental in collecting high-precision survey-grade point cloud data. By leveraging such data, DOTs can automate asset management, monitor infrastructure health, and ensure compliance with standards such as the ADA. This reflects a broader trend towards digital transformation in infrastructure management.

\begin{figure}[!t]
\centering
\includegraphics[width=\linewidth]{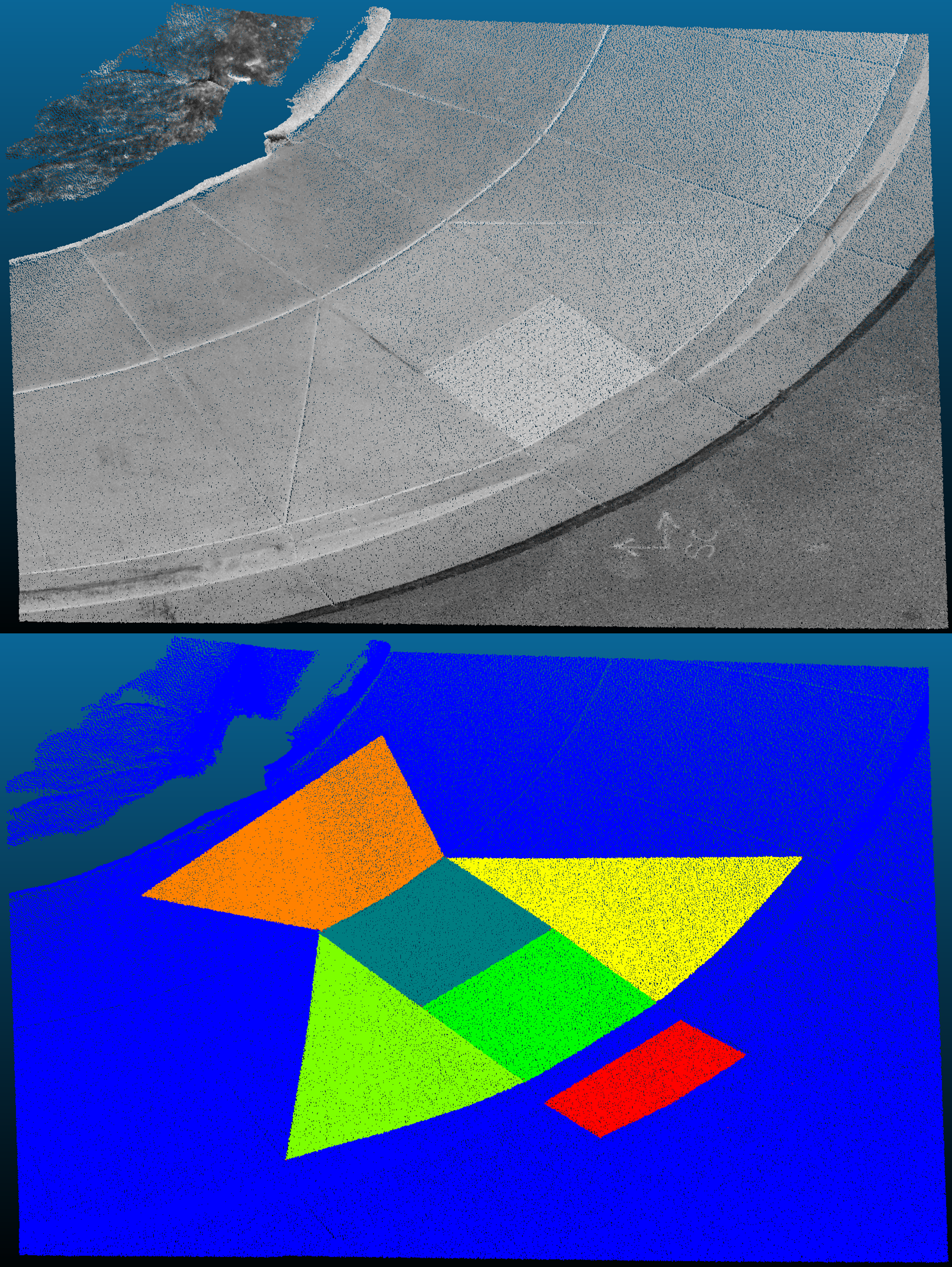}
\caption{An example ramp point cloud (top) and its manual segment annotations (bottom). Due to the ambiguity in the landing area for visual annotation, this segment (orange) is annotated with expansion going away from the ramp as long as the visually flat sidewalk area would allow. This is to ensure the annotation incorporates the correct landing area.}
\label{fig: sample_ramp_segmentation}
\end{figure}

In this research, we utilize MTLS data collected by Caltrans from various counties in Northern California. While these datasets provide a comprehensive representation of urban infrastructure, they lack the annotations necessary for tasks such as ramp detection and segmentation. To address this, we manually annotated a sizable portion of the data, focusing on ramps and their key components. Figure~\ref{fig: sample_ramp_segmentation} illustrates an example of a segmented ramp from the dataset. The annotated dataset, which we make publicly available, includes over 1600 segmented ramps.

Following the proposed framework, we next detail the process for automated ADA compliance assessment of curb ramps.

\subsection{Asset (Ramp) Detection}
This section illustrates the ramp detection method (step 2 in Fig. \ref{fig: flowchart}), which aims to localize and extract individual ramps from point cloud data. 

\subsubsection{Comparison of 3D and 2D Object Detection}
For ramp detection, both 3D and 2D object detection techniques were considered. 3D object detection directly processes the point cloud data, leveraging the full spatial structure of the environment. While this method generally captures detailed spatial information, it usually requires significantly higher computational resources and complex models, making it less efficient for large scale processing. In addition, ramps often have minimal distinct features compared to the surroundings such as the adjacent sidewalks. This lack of clear structural distinction makes it challenging for 3D object detection models to reliably isolate ramps. 

On the other hand, the problem is simplified by projecting the 3D point cloud onto a 2D horizontal plane after elevation thresholding. In 2D, we can make use of well-established and robust 2D object detection models and reduce computational complexity. The normalized laser intensity values help compensate for the loss of spatial information during top-down projection by enhancing edge and texture delineation. As such, we choose 2D object detection for ramp extraction. 

\subsubsection{Point Cloud Preprocessing for Ramp Detection}
\begin{figure}[!t]
\centering
\includegraphics[width=\linewidth]{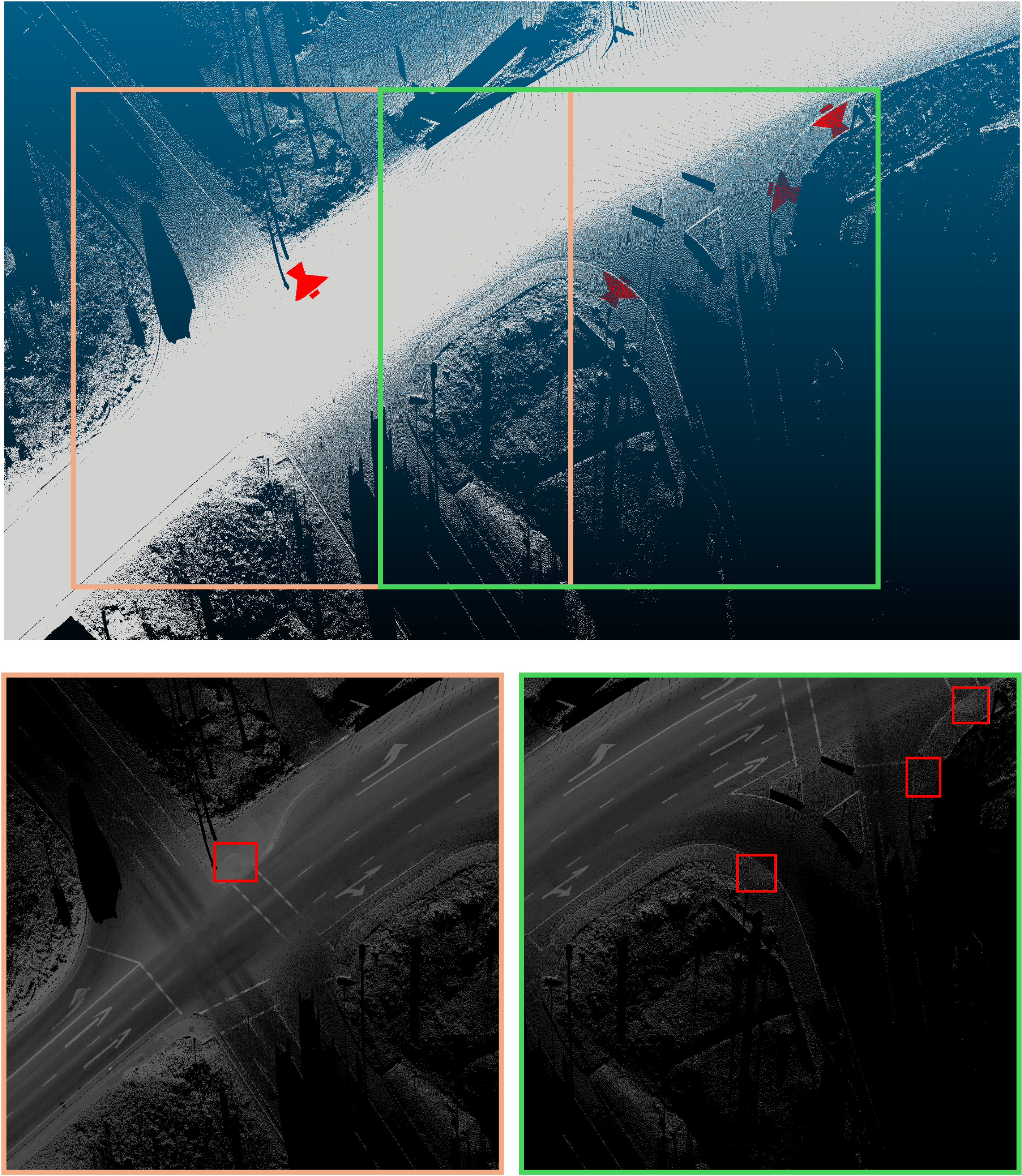}
\caption{Top: partial street-level point cloud with segmentation masks (shown in red) projected onto the ground plane. Example overlapping patches are outlined in orange and green. Bottom: corresponding patches where 2D segmentation masks have been converted into tight bounding box annotations (shown in red).}
\label{fig: ramp_dataset_prep}
\end{figure}

To prepare the dataset for training ramp detection models, our segmentation annotations are first converted into bounding box annotations. The street-level point cloud with segmentation masks is projected onto the ground plane and divided into multiple overlapping square patches (top image in Fig. \ref{fig: ramp_dataset_prep}). Each patch is rendered on a $1280 \times 1280$ canvas. We use the projected segmentation masks to construct a tight bounding box around each ramp (bottom images in Fig. \ref{fig: ramp_dataset_prep}). In this step, bounding box coordinates are saved separately as labels in Microsoft COCO (MS-COCO)\cite{lin2014microsoft} format, and intensity values are extracted and normalized to a 255 scale, which are represented as grayscale values. All transformations are saved for each patch to ensure that the 2D projection can be accurately converted back to 3D space when needed. 

This preprocessing step results in 2D grayscale images with 2D bounding box annotations, which we will refer to as the ramp detection dataset in the rest of the document.

\subsubsection{Object Detection Models}
We compared Faster R-CNN \cite{ren2016faster} and DeTR \cite{carion2020end} for ramp detection, and chose DeTR as our final object detection model. While Faster R-CNN is effective for general object detection, in our application it struggles with small or indistinct features, which are critical for isolating ramps from the surroundings. DeTR employs a transformer-based architecture that predicts object positions and classes in an end-to-end manner. Its global attention mechanism enables it to capture broader contextual cues across the image, making it more suitable in scenarios where ramps exhibit minimal visual distinction from their surroundings. 

In the comparison of the two models, we fine-tuned their pre-trained weights on our ramp detection dataset. We evaluated both ResNet-50 and ResNet-101 backbones and selected the latter for our DeTR model due to its superior performance.

\subsubsection{Data Preprocessing for Inference}
During deployment on unseen data, we follow the same preprocessing steps used during training. The point cloud is divided into overlapping square patches and projected top-down onto a $1280\times1280$ canvas. Intensity values are normalized to grayscale, and the transformation metadata necessary for mapping predictions back to 3D space is preserved. These preprocessed 2D images are then fed into the trained ramp detection model, which outputs 2D bounding boxes for the ramp class.

\begin{figure}[!t]
\centering
\includegraphics[width=\linewidth]{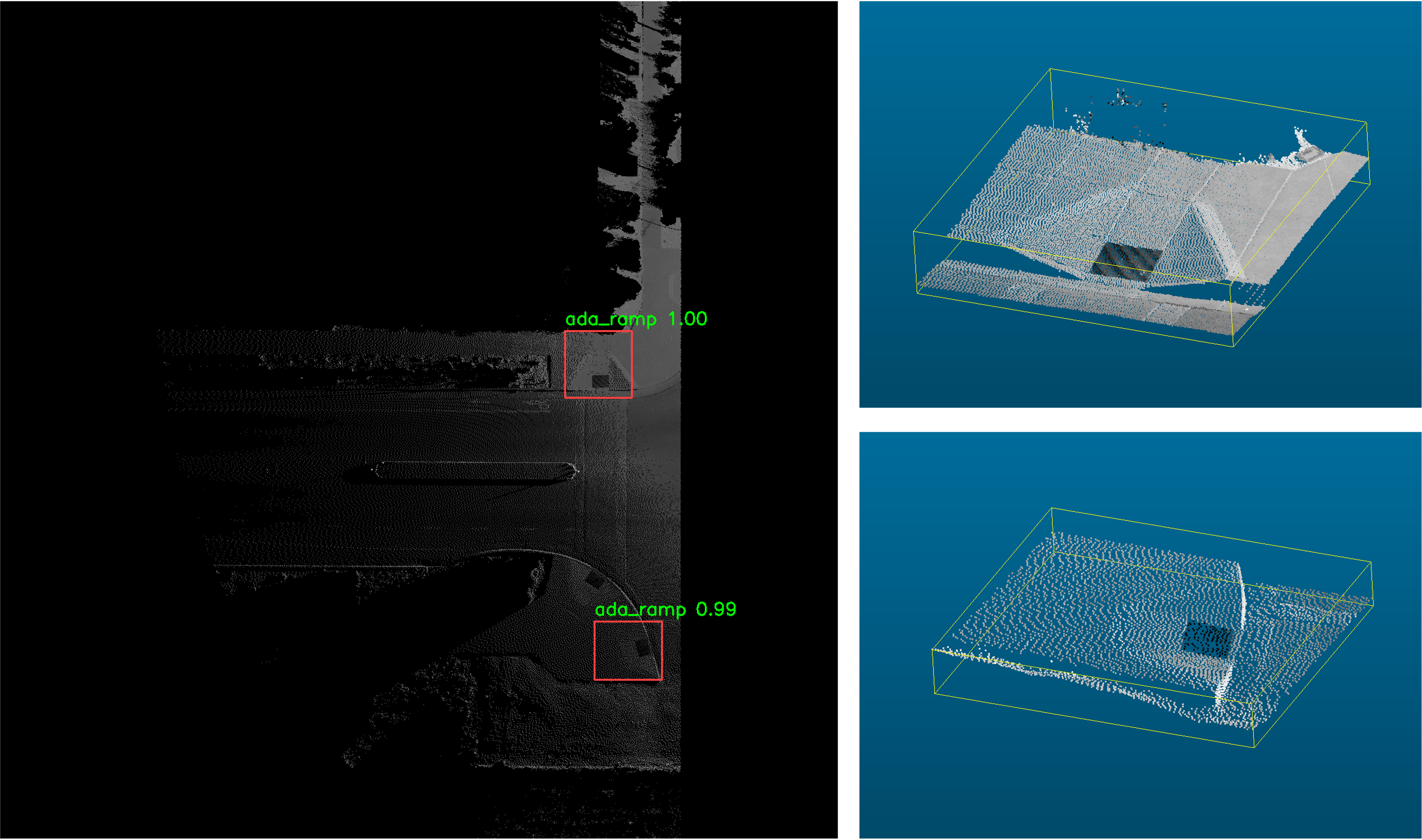}
\caption{Left: 2D image with detected ADA ramps highlighted with predicted 2D bounding boxes and confidence scores. Right: corresponding 3D crops of the original point cloud, reconstructed from the 2D predictions using stored transformation metadata.}
\label{fig: ramp_detection}
\end{figure}

\subsubsection{Extracting Individual Ramps}
As a result of ramp detection, bounding boxes are predicted for each 2D image (left image in Fig. \ref{fig: ramp_detection}). Using the transformation metadata, the 2D image and the corresponding bounding box coordinates are converted back to 3D space. The predicted 3D bounding boxes are then mapped back to the original point cloud (right images in Fig. \ref{fig: ramp_detection}). Each bounding box is extended by moving its edges outwards by 10\% of their original length to include any adjacent points that may belong to the ramp but fall outside the detected boundary. Finally, the extracted 3D ramps are saved in separate files for downstream segmentation and measurement.

\subsection{Visual Segmentation of Detected Ramps}

Here we present our approach for visual ramp segmentation, i.e. step 4 of the block diagram of Fig. \ref{fig: flowchart}. This process is applied to the ramps detected through the previous step. 
This segmentation step is challenging due to (a) limited training data, (b) the high dimensionality of 3D point clouds, and (c) non-uniform point density, requiring robustness to point clouds of varying quality and density.

To address (a) and (b), we opt for segmentation in the 2D image space. For this purpose, we use the Segment Anything Model (SAM)~\cite{Kirillov_2023_ICCV}, a strong pretrained image segmentation model that can efficiently adapt to custom tasks with minimal training data. For effective 2D representation and to mitigate the effect of density variation as specified in (c), we incorporate additional steps for preprocessing, which we describe in the following.

\subsubsection{Point Cloud Preprocessing for Ramp Segmentation}
Like before, to render the point cloud as a 2D image, it is projected down to the horizontal plane and visualized as a grayscale image---note that even though in both this and the previous step the point clouds are mapped to 2D images, at each stage we present the output as the full 3D point cloud for more flexibility and modularity in designing different stages of the automation. The projection, however, generates a sparsely colored image with varying local densities reflecting the local point cloud density, resulting in poor segmentation performance. As such, we apply pixel dilation to fill out the empty areas between the pixels and form contiguous regions.

Dilation, however, faces additional challenges. Notably, varying point cloud densities prevents adoption of a global scale value for dilation. If we consider a small kernel, it will miss significant empty areas, leaving the lower-density areas still highly sparse and discontinuous. On the other hand, an unnecessarily large kernel can obscure detailed ramp information and importantly, ramp boundaries.

\begin{figure}[!t]
\centering
\includegraphics[width=\linewidth]{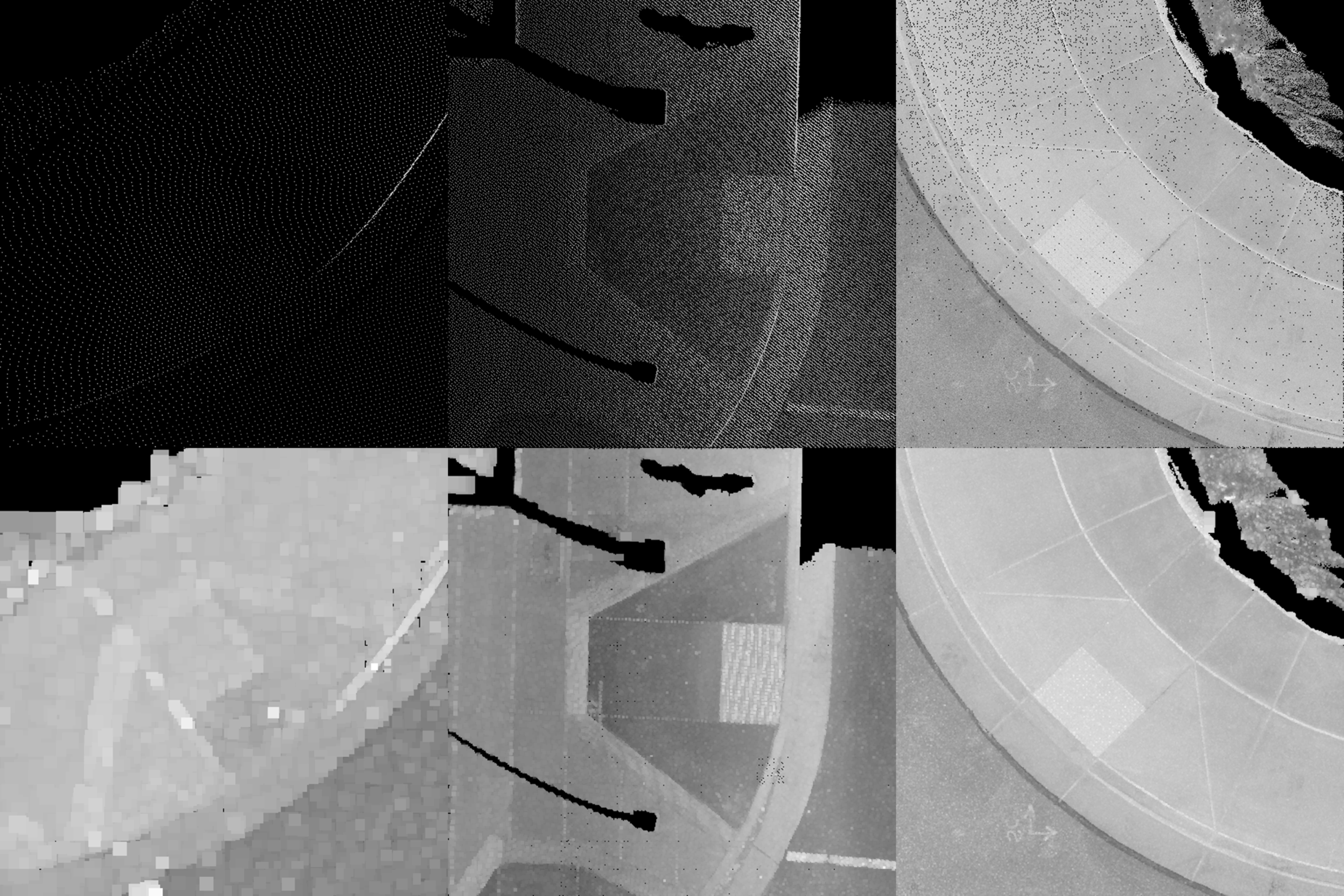}
\caption{Grayscale images associated with the top-down view of point cloud with varying local densities before (top) and after (bottom) applying adaptive dilation. Adaptive dilation improves the image contiguity and hence, the segmentation performance.}
\label{fig: dilation}
\end{figure}

To address this, we propose an adaptive dilation approach, which adjusts the dilation kernel size to the local density, improving contiguity while preserving detail. The process entails tiling the ramp images, computing the local density for each patch, and subsequently determining and applying an optimal kernel size for each. The square dilation kernel size $\kappa$ for the patch is calculated using
\begin{align}
\kappa = \min\left(\left\lfloor \frac{1}{2\rho_{proj}} \right\rfloor, \kappa_{max}\right),
\label{eq: kappa}
\end{align}
where $\rho_{proj}$ is the projected point cloud density in the image domain and is derived by dividing the count of non-zero-intensity pixels by the total number of pixels in that patch. We set an upper bound for $\kappa_{max}$ (a hyperparameter) to prevent the kernel from becoming too large for extremely sparse patches. Subsequent to the local dilation step, a global dilation with a kernel size of 2 is then applied to ensure image contiguity. Figure~\ref{fig: dilation} shows examples before and after adaptive dilation.

The ground-truth segmentation masks are similarly treated as grayscale images. Pixel intensity values are used to represent the associated class in a ground-truth mask. Once the point cloud data is processed to create the 2D images and their associated segmentation masks, they are used to finetune the SAM model.

\subsubsection{Ramp Segmentation with SAM}
SAM model consists of three main parts: an image encoder, a prompt encoder, and a lightweight mask decoder. The role of the prompt encoder is to receive any available region proposals and cues in the form of either a bounding box, prior mask, or text prompt. For our use case, the prompt is provided as a bounding box that contains the entire image. We modify the architecture so that the same image and prompt encodings are fed to multiple independent mask decoder modules, each specializing in an independent segmentation class.

\subsection{Leveraging Geometric Primitives}

The ramp segments recovered so far are purely based on the visual cues, and in particular, those present in the 2D top-down view. In this section, we describe how we can incorporate geometric priors, which is especially suited here because they can be expressed rather simply given that the ADA ramp is a man-made structure. The priors match the intended design of the ADA ramp. For instance, each component surface is assumed to be flat, featuring a pre-defined arrangement of its corner points. Naturally emerging from this, adjacent component interfaces should be well-approximated as a line. In the following steps, we leverage these geometric primitives in developing our hybrid automation algorithm.

\subsubsection{Ramp Decomposition into Constituting Components}
\begin{figure*}[ht!]
    \centering
    \subfloat[]{%
        \includegraphics[width=0.33\textwidth]{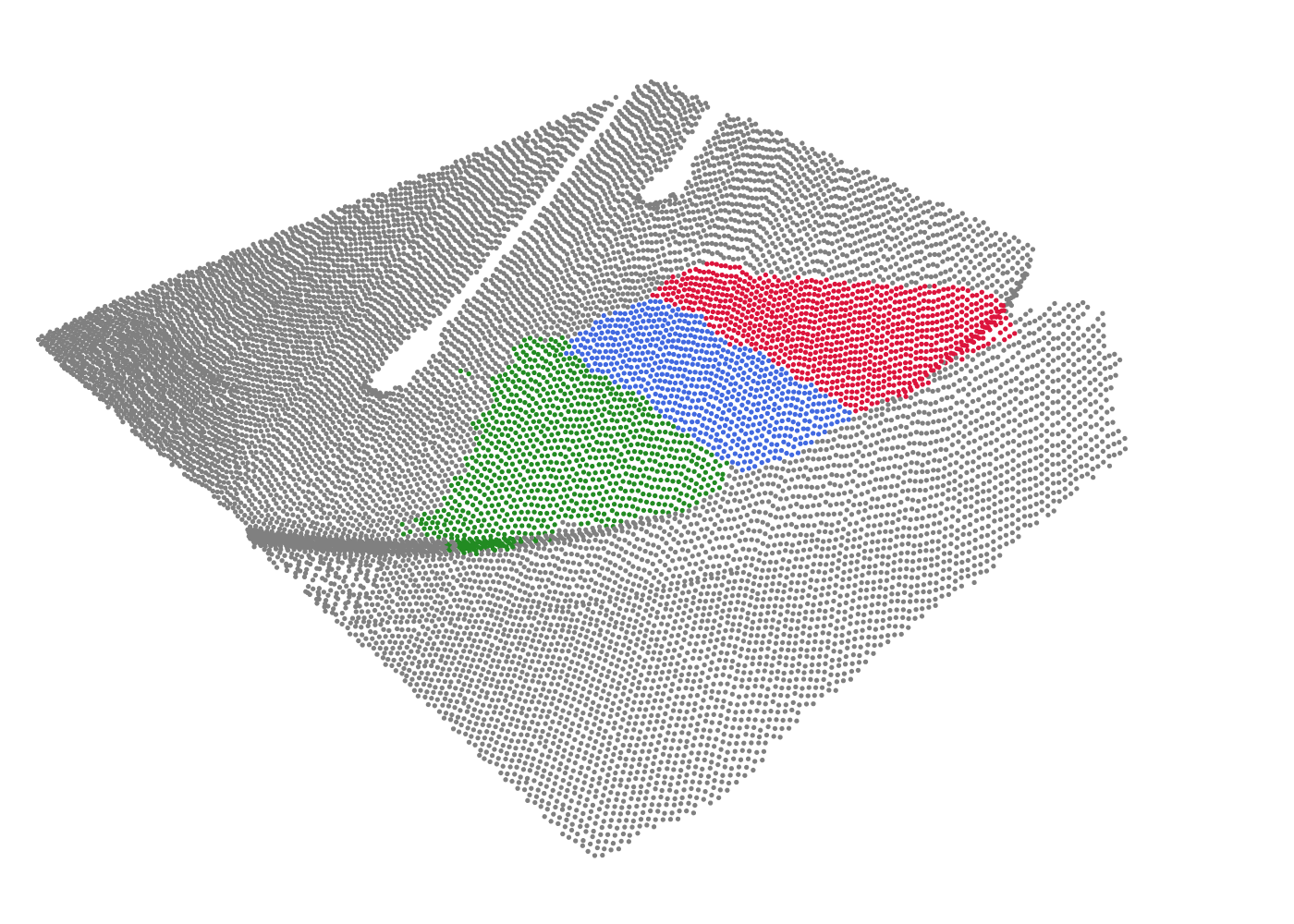}%
        \label{fig:input}}%
    \hfill
    \subfloat[]{%
        \includegraphics[width=0.33\textwidth]{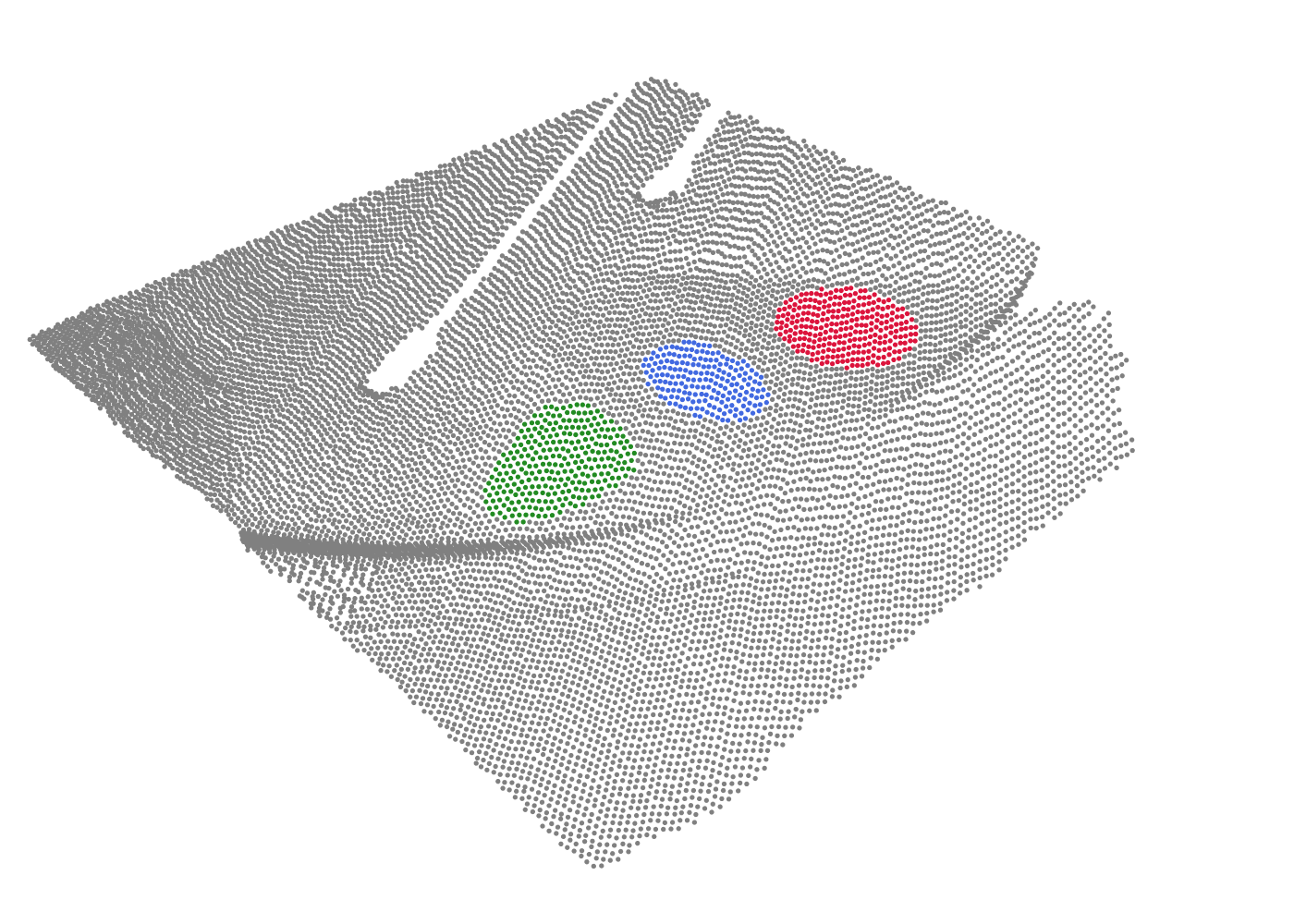}%
        \label{fig:ocsvm}}%
    \hfill
    \subfloat[]{%
        \includegraphics[width=0.33\textwidth]{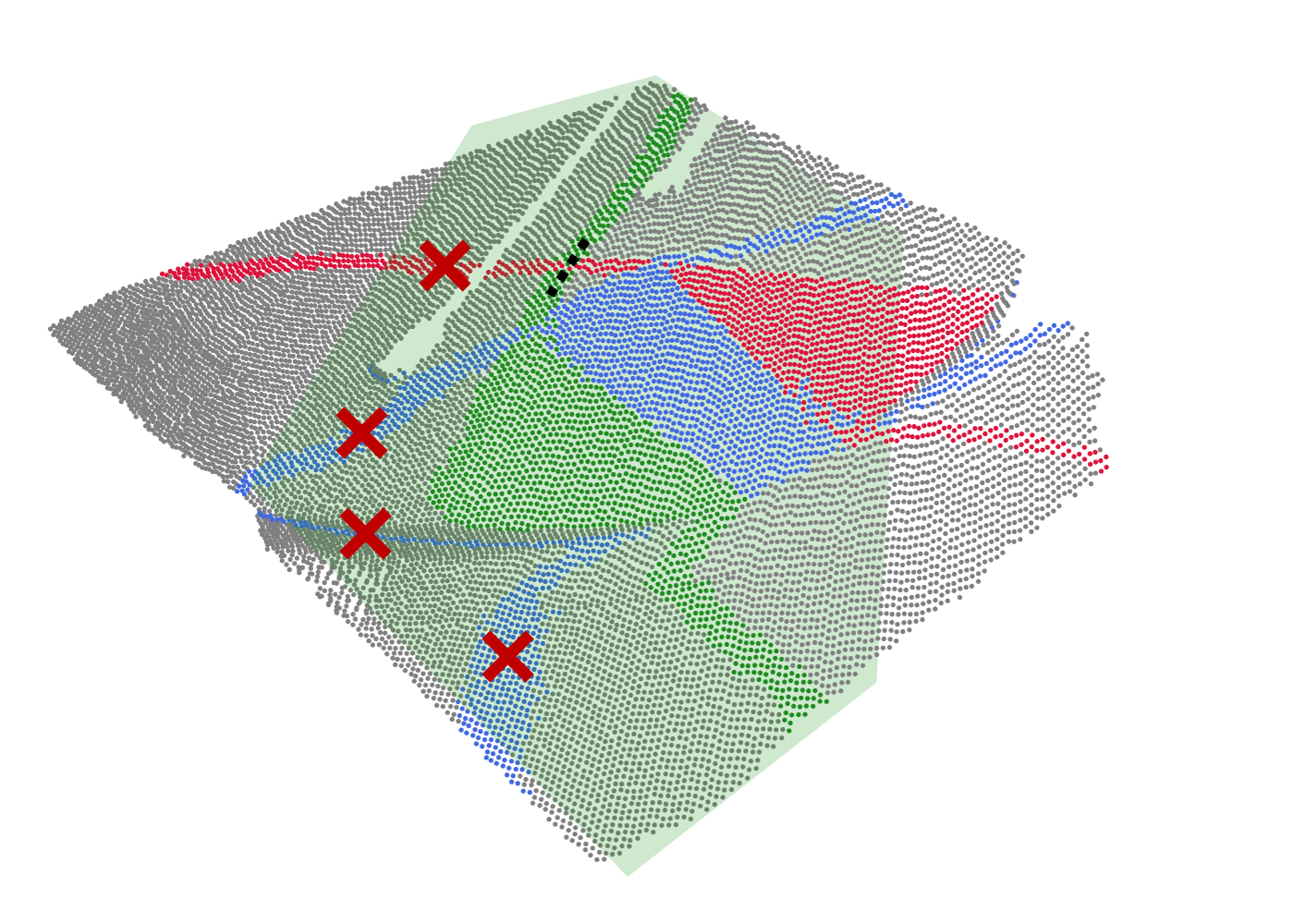}%
        \label{fig:planefitting}}%
    
    \vspace{1em} % Space between rows
    
    \subfloat[]{%
        \includegraphics[width=0.33\textwidth]{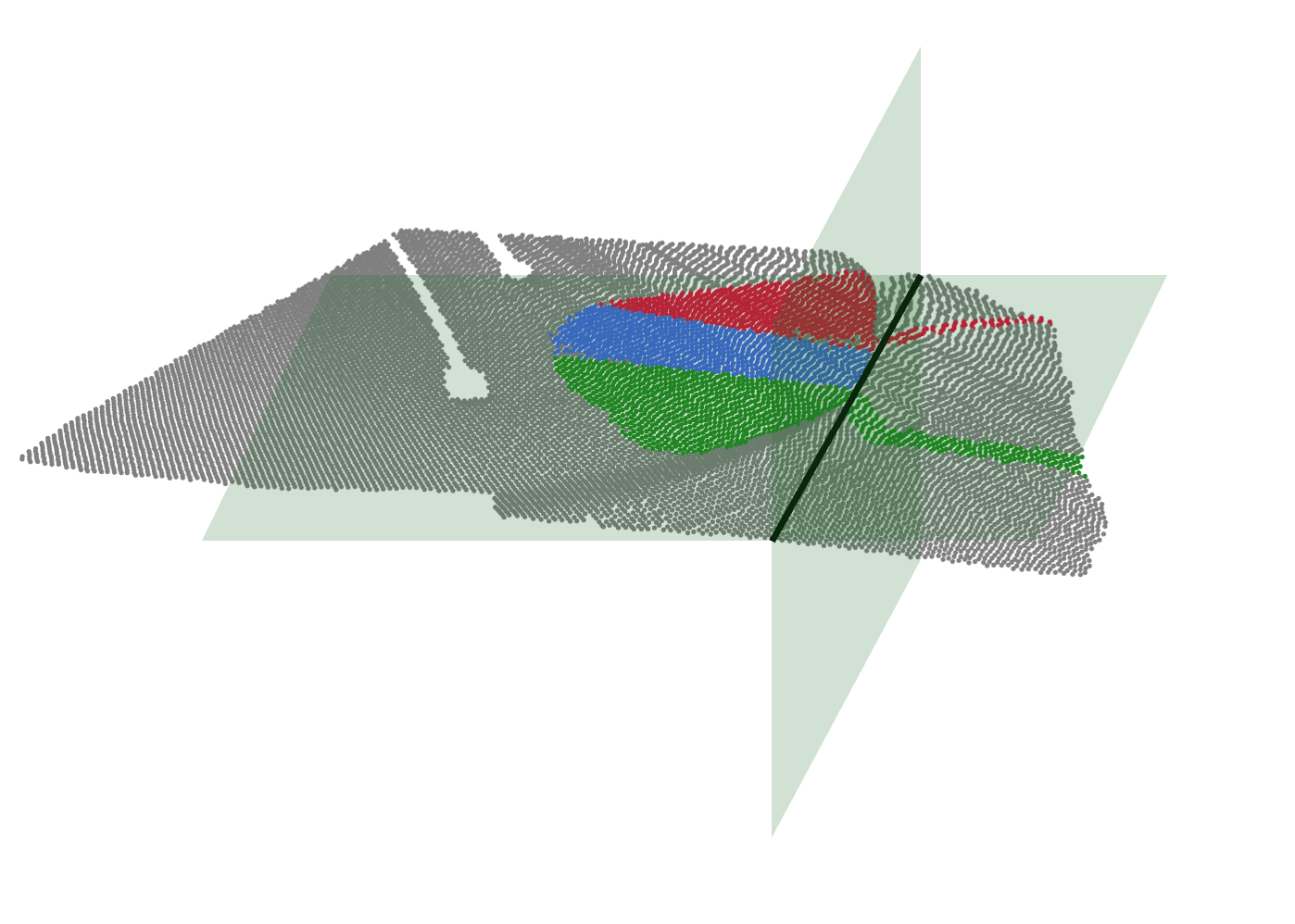}%
        \label{fig:quadrant}}%
    \hfill
    \subfloat[]{%
        \includegraphics[width=0.33\textwidth]{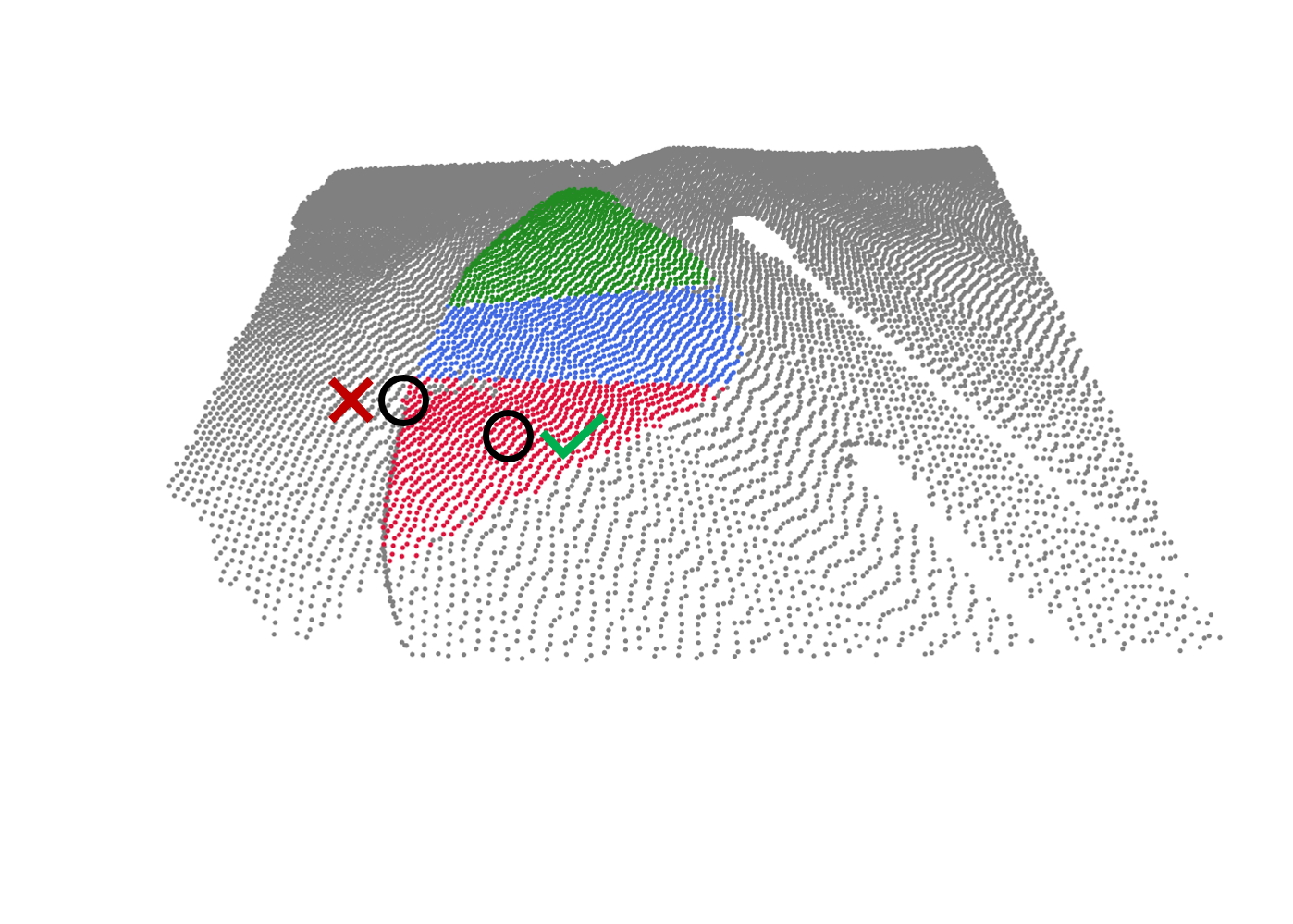}%
        \label{fig:simfiltering}}%
    \hfill
    \subfloat[]{%
        \includegraphics[width=0.33\textwidth]{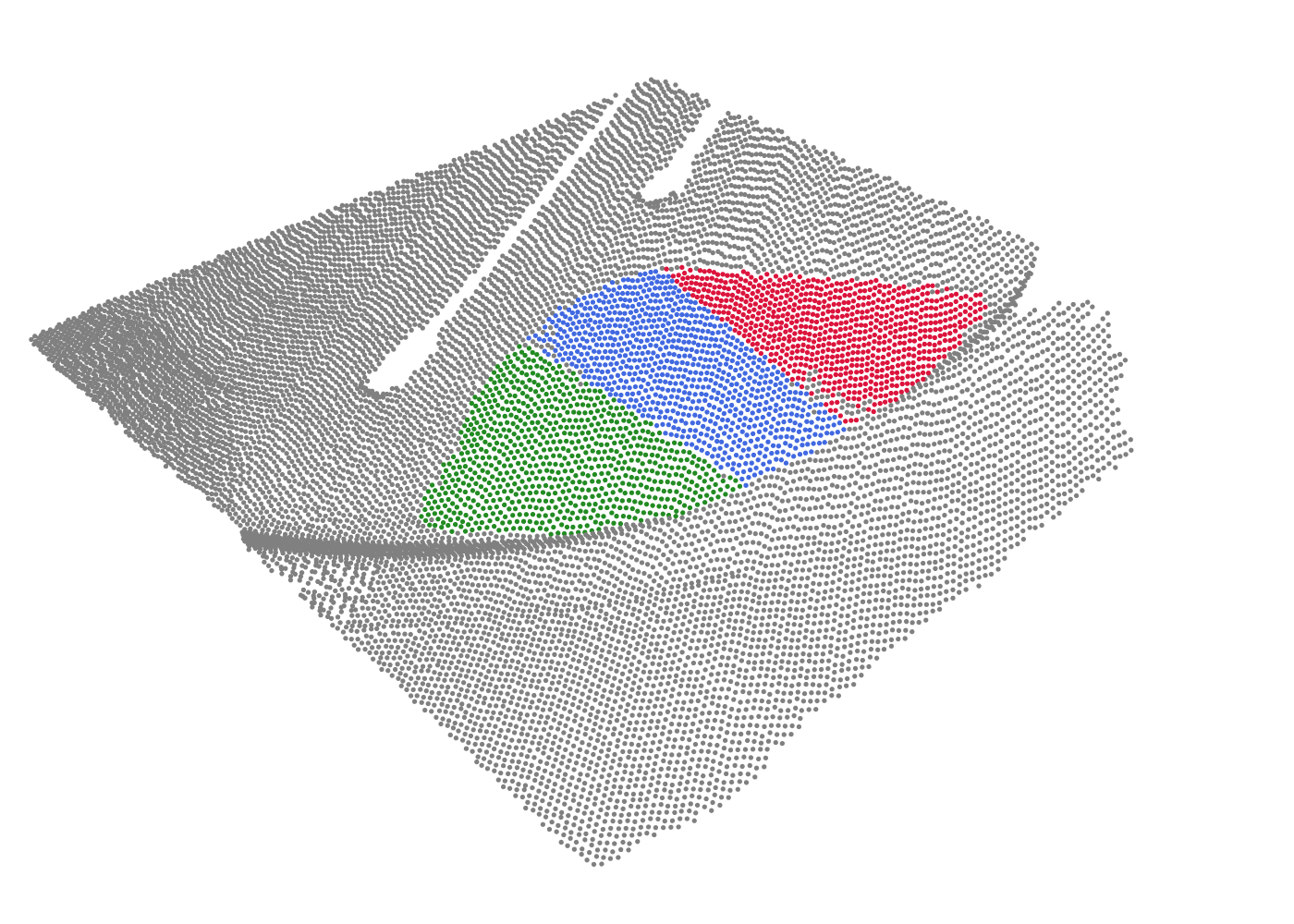}%
        \label{fig:output}}%

    \caption{Geometric component decomposition. (a) Visual segmentation results, which are generally crude, agnostic to the vertical coordinate, and based solely on visual features in the associated top-down 2D image. (b) More reliable internal points for each segment after applying OCSVM. (c) Re-classification based on proximity to the internal coreset plane fits. The plane fit for the green component is shown. Re-classification based on proximity alone leads to artifacts, primarily in ribbon-like patterns, a result of plane intersection with unassigned parts of the point cloud. The three major ramp planes are used to cross-filter some of these artifacts. For example, red crosses denote ribbons associated with other components that can be cleaned out using the green plane. For the red ribbon, only the portion beyond the black dashed line is cleaned when filtering via the green plane. The process is repeated for all planes, where incorrect labels are removed from the wrong half-space for each plane. (d) Using the bottom line fit to the center component, the vertical and the corresponding perpendicular planes passing through the line are used to filter out any component labels from the three incorrect spatial quadrants, leaving component labels only in the correct quadrant (top left). (e) For each labeled point, its local plane fit is compared with the component plane fit. If the absolute cosine similarity between their normal vectors falls below a critical threshold, the point will be removed from the component. Two neighborhoods associated with a correct and an incorrect point are demonstrated respectively with a green check and a red cross next to them. (f) Final decomposition output after iForest anomaly detection and final DBSCAN.}
    \label{fig:mainfigure}
\end{figure*}

\begin{figure}[ht!]
\centering
\includegraphics[width=\linewidth]{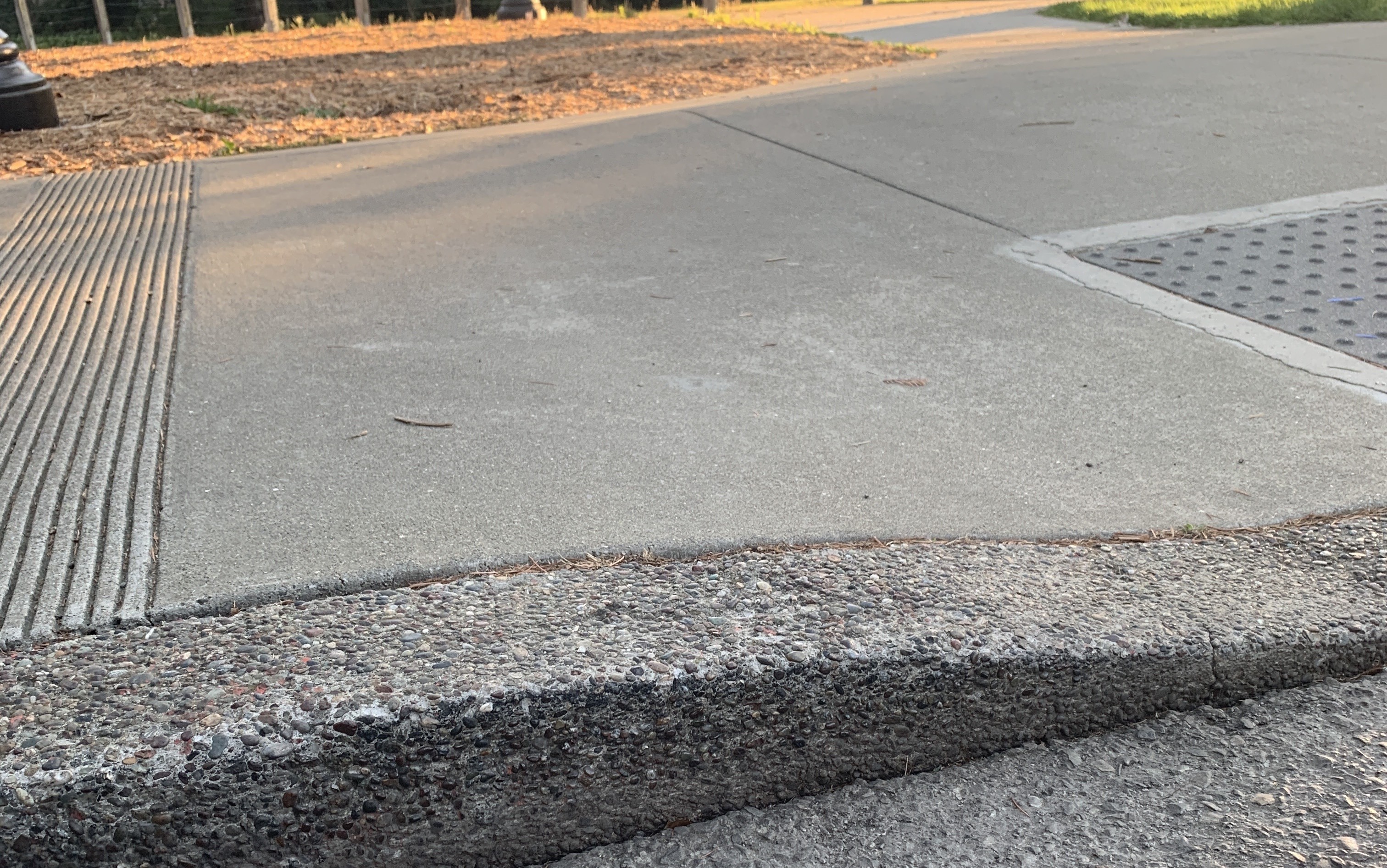}
\caption{Photo of a side flare. The visual features often do not overlap with geometric attributes and the deviations can be significant.}
\label{fig: aesthetic}
\end{figure}

In this section we discuss step 5 of the block diagram of Fig. \ref{fig: flowchart}. The segmentation results obtained from the previous step are converted back to 3D along with their identified class labels (Fig.~\ref{fig:input}). These segments reflect an approximate separation of ramp components purely based on visuals such as the concrete joints. Ideally, however, the separation should correspond to the geometric structure, i.e. the constituting planes, rather than visual segments. In practice, there is often no such one to one match between the geometric separation and said aesthetics. In fact, deviations between the two are often significant (Fig.~\ref{fig: aesthetic}), even though surveying guidelines inherently assume that visual and geometric boundaries align. Furthermore, as a result of performing 2D image segmentation, there can be points that are vertically separated but overlapping when projected on the horizontal plane and thus incorrectly assigned to a particular segment. The limitations of segmentation combined with the noted distinction between visual versus geometric components motivate further refinement of the segments before carrying out the measurements.

A key observation is that while the segmentation result does not adequately serve as the component separation endpoint that is suitable for measurement, it still provides us with a valuable prior. More specifically, we observe that the segmentation model almost always retrieves segments with correct internal points even for very poor-quality outputs. We base the geometric refinement on this premise, namely the inliers associated with a visual segment are assumed to overwhelmingly belong to the correct geomtric component. The second premise comes from the prototypical design of ADA ramps, which specifies standards for surface flatness, hence requiring each segment to be well-approximated by a plane, with the possible exception of the gutter section.

We restrict our geometric refinement step to the center ramp and warning surface, as well as the side flares. Later in the text we will describe how the landing and gutter areas are handled. Further, considering that the warning surface and the center ramp should be on the same plane, we combine these two components in our analysis when the warning surface is present. This means that we strive to identify and separate 3 constitutive components: center ramp plus warning surface if present (parts 1 and 2 in Fig. \ref{fig: ramp_components}), left flare, and right flare (parts 3 and 4 in Fig. \ref{fig: ramp_components}). Importantly, we seek a method that does not rely on manual intervention and customization for different ADA ramps.

\paragraph{Plane-Based Reassignment}
We begin by removing the noisy points of each of the three components as initially provided by the visual segmentation. To do this, we carry out an aggressive outlier detection and removal for each component. We employ one-class support vector machines (OCSVM)~\cite{ocsvm} with a radial basis function (RBF) kernel $K(x, y)=\exp{\left( -\gamma||x-y||^2\right)}$ and a slack variable $\nu$ for this purpose. $\nu$ can be interpreted as an upper bound on the fraction of outliers in each component.

Once a coreset is retrieved for each component (Fig.~\ref{fig:ocsvm}), it is used to fit a plane. Following the plane fits, each point is assigned to the closest plane if its distance to that plane is less than a predefined threshold $t$. As a result, the ramp and flare boundaries now more accurately reflect the geometric surface characteristics. This step also removes the initially misclassified vertically separated points.

We note that depending on the threshold $t$, we might also remove points from each component that should nominally belong to it. However, we emphasize that the requirement for this stage is to have labels that can assist with defining references such as corners and boundaries that are useful in measurement extraction, rather than correct class assignment for every point. During the measurement step, the potentially missed points will still contribute to the calculations whose values are partly affected by such points.

\begin{algorithm}[!t]
\caption{Geometry-Informed Component Refinement}
\label{alg:component_refinement}
\begin{algorithmic}[1]
\REQUIRE \textbf{Segments:} $\{S_i\}$, where $S_i$ corresponds to center ramp and side flares \\
         \textbf{Point clouds:} $\{P_i\},U$ where $P_i$ are points assigned to $S_i$ and $U$ is the unassigned point set\\
         \textbf{Parameters:} OCSVM slack variable $\nu$, RBF kernel coefficient $\gamma$, plane distance threshold $t$, local neighborhood ball radius $r$, cosine similarity threshold $s_{crit}$, iForest contamination and number of trees, DBSCAN distance parameter $\varepsilon$
\vspace{4pt}
\STATE \textbf{(A) Extract Internal Coreset:} 
\FOR{each component $S_i$}
  \STATE Identify the coreset $C_i \subset P_i$ via OCSVM; 
\ENDFOR
\vspace{4pt}
\STATE \textbf{(B) Plane Fits and Reassignment:}
\FOR{each $C_i$}
  \STATE Fit a plane $\pi_i$ to $C_i$;
\ENDFOR
\FOR{each $p \in \bigcup_i P_i \;\cup\; U$}
  \STATE Find plane $\pi_j$ with $\min\text{dist}(p,\pi_j)$;
  \IF{$\min\text{dist}(p,\pi_j) < t$}
    \STATE Assign $p$ to component $j$;
  \ELSE 
    \STATE Unassign $p$ if already assigned;
  \ENDIF
\ENDFOR
\vspace{4pt}
\STATE \textbf{(C) Cross-Filtering via Half-Spaces:}
\FOR{each $p \in P_i$ and plane $\pi_j,\quad i\neq j$}
  \STATE Unassign $p$ if it lies in the wrong half-space w.r.t. $\pi_j$;
\ENDFOR
\vspace{4pt}
\STATE \textbf{(D) Filtering via Quadrant Check:}
\STATE Identify the bottom edge of $P_k$ through SBLF ($k$ is the index corresponding to center ramp);
\STATE Fit a vertical plane $\pi_{b1}$ and its perpendicular plane $\pi_{b2}$ passing through the bottom edge;
\FOR{each $p \in \bigcup_i P_i$}
  \STATE Unassign $p$ if it lies in a disallowed quadrant w.r.t. $\pi_{b1}$,  $\pi_{b2}$;
\ENDFOR
\vspace{4pt}
\STATE \textbf{(E) Local Normal Consistency:}
\FOR{each point $p \in P_i$}
  \STATE Collect neighbors $N_p$ of $p$ within radius $r$, biased toward the centroid of $C_i$;
  \STATE Fit local plane $\pi_{p}$ to $N_p$;
  \IF{$|\text{cosineSim}(\text{normal}(\pi_{p}), \text{normal}(\pi_{i}))| < s_{crit}$}
    \STATE Unassign $p$;
  \ENDIF
\ENDFOR
\vspace{4pt}
\STATE \textbf{(F) Outlier Removal:}
\STATE Apply iForest and DBSCAN to remove sparse outliers;
\RETURN Refined assignments $\{P_i^\ast\},U^\ast$ 
\end{algorithmic}
\end{algorithm}

\paragraph{Clean-Up}
A consequence of reassigning points solely based on plane fits is that it often leads to formation of artifacts, external to the boundaries of ramp components, resulting from plane fits crossing through the entire point cloud, and capturing many points on the sidewalk and street/gutter. These artifacts include continuous, typically ribbon-like extensions above, below, and to the sides of the ramp area (Fig.~\ref{fig:planefitting} areas indicated by red cross). While anomaly detection and clustering algorithms like DBSCAN~\cite{dbscan} can help, we find it difficult, if not impossible, to pin down a set of universal hyperparameters that lead to robust performance across all or even a large fraction of ramps, making the choice impractical. This is due to the high variability across ramps and their surroundings. Instead, in the following we describe a primarily rule-based approach with minimal hyperparameters that are much easier to set for robust performance across different ramps.

First, relying on our a priori knowledge of the ramp construct, we highlight that at the primitive geometric level, an ADA center ramp and side flares constitute a (piecewise flat) concave surface by design. For components $i$ and $j$ of such an idealized surface, all points belonging to component $i$ are only on one side of component $j$ and vice versa (Fig.~\ref{fig:planefitting}). Applying this observation to the assignments, we check whether each point assigned to component $i$ is located in the correct half-space with respect to the plane fit for components $j\neq i$. To determine if a point is in the correct half-space relative to component $j$, we project it onto the normal vector of the plane fit associated with component $j$. We measure this projection with respect to the initial coreset centroid of this component. We then compare the sign of this projection to that of the initial coreset centroid of component $i$, projected along the same normal vector. If both projections have the same sign, the point is on the correct side of the plane. We repeat this cross-check for all six possible combinations of components and each time remove the label of points that reside on the wrong side of the plane.

This approach removes mislabeled points and ribbons on the sidewalk and the curbs. However, ribbons often misclassified as side flares remain on the street side.

To correct the remaining misclassifications, we fit a line to the bottom boundary of the center ramp component. The details for the bottom line fitting will be described later in Sections~\ref{linefitting} and~\ref{linefitting2}. To this line, we first fit a vertical plane and then a second plane perpendicular to this vertical plane. For each plane, we remove mislabeled points from the wrong half-space with the ultimate goal of only keeping assigned points in the correct quadrant (see Fig. \ref{fig:quadrant}). Like before, the correct half-space is determined by a sign match between the out-of-plane coordinate of a point and the center ramp coreset centroid with respect to the plane’s coordinate system (Fig.~\ref{fig:quadrant}). While this step removes the majority of remaining misclassifications, it does not entirely safeguard measurement accuracy.

We note that at this stage the remaining misclassifications reside on surfaces with local plane fits that are different from their assigned component plane fit. We can leverage this mismatch to further purify the points. To do this for each component, we first obtain the nearest neighbors of each point (Fig.~\ref{fig:simfiltering}). The neighborhood distance $r$ should be selected to accommodate varying point cloud densities, and to ensure that the remaining misclassified points do not dominate the neighborhood. For each point, we can then obtain a plane fit to its local neighborhood. If the absolute value of cosine similarity between a local normal vector and the component normal vector fall below a defined threshold $s_{crit}$, the point can be dismissed as a misclassification. A caveat here is that for the correctly classified points close to the boundaries of the center ramp or flares, a local neighborhood will contain a significant number of points from the adjacent component and/or non-ramp areas. To mitigate this, we can bias the local neighborhood such that only neighbors to the correct side of a point are considered. Empirically, we define the correct side of the point as one which is closer or equal in distance from the initial component coreset centroid compared to the point. Biasing the neighborhood in this way leads to a neighborhood of completely or overwhelmingly correct points, leading to a local fit that closely resembles the component plane fit.

By this stage, we are typically left with sparse outliers which are easily removed through Isolation Forest (iForest)~\cite{iforest}. While the optimal value for the contamination parameter (i.e. the expected proportion of outliers) depends on the specific ramp, we find that the search region for this parameter is relatively small and that a fixed small value can yield an acceptable performance across different ramps. 

Finally, we apply DBSCAN to the set of all labeled points in case after the anomaly detection there still remain isolated incorrect points that this step can correct. We remove the label of any points not belonging to the largest cluster, which corresponds to the combined center ramp and flares. An example output for this process is shown in Fig.~\ref{fig:output}. Algorithm~\ref{alg:component_refinement} summarizes the steps to refine visual segmentation results.

\subsubsection{Reference Point Detection}
\begin{figure}[!t]
\centering
\includegraphics[width=\linewidth]{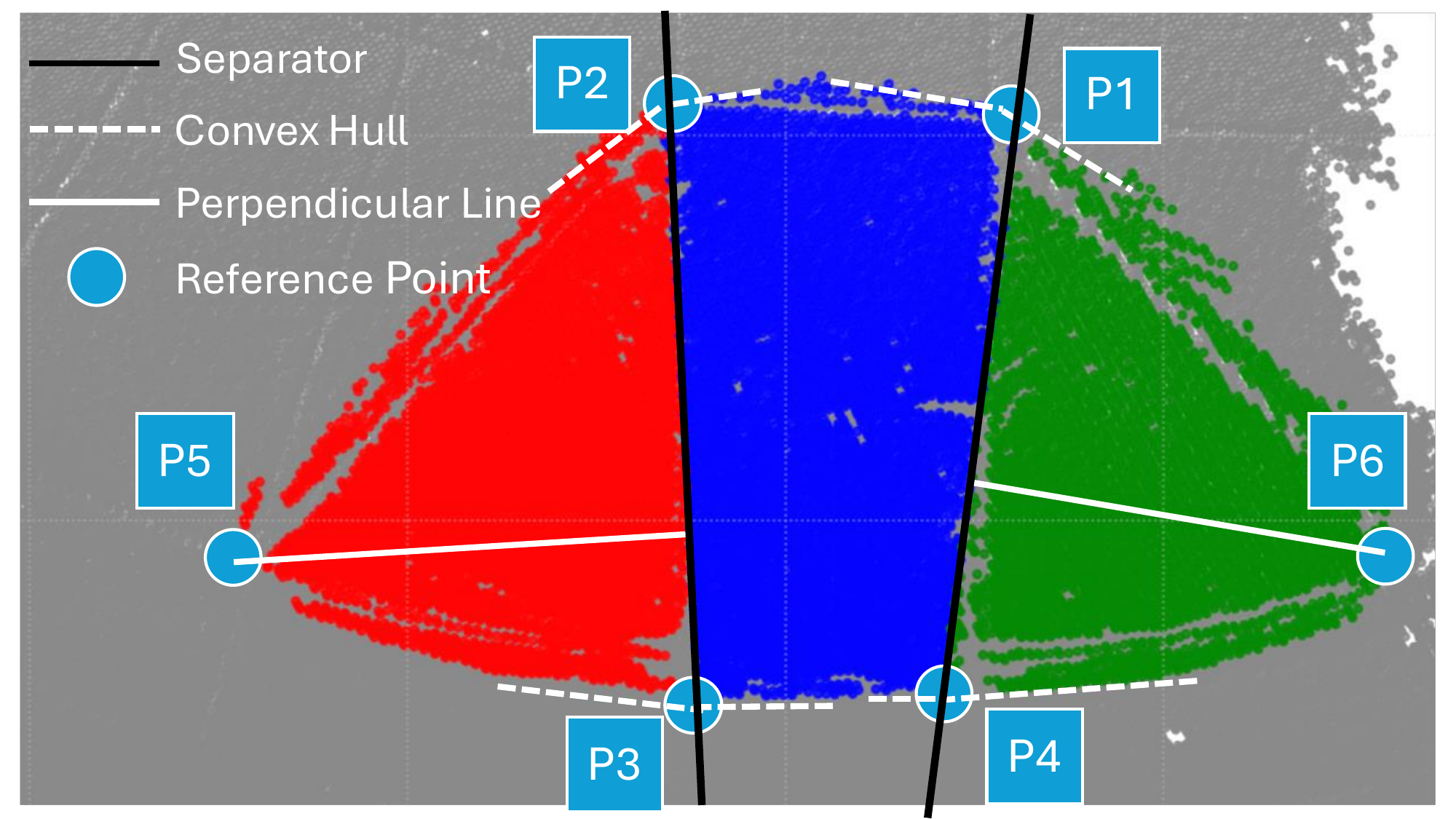}
\caption{Our approach extracts reference points through a multi-stage process. By using bird's eye view of the ramp point cloud, we enable efficient 2D processing while preserving geometric relationships. We employ our novel Score-Based Line Fitting (SBLF) algorithm to identify separators (black) between center ramp and two flares. The four corners (P1-P4) on the center ramp are given by the intersections between the separators and the ramp convex hull (dashed white lines). Finally, we leverage structural priors to detect two flare corner points (P5 and P6) by identifying locations with maximum projection distance to two separators (white solid lines).}
\label{fig: ref_point}
\end{figure}

Detection of reliable reference points is crucial for accurate geometric measurements of curb ramps. However, despite the processing steps taken so far, the data can still contain noise and outliers, and naturally suffer from varying density. To address these challenges, wedevise a robust reference point detector (step 6 in the block diagram of Fig.~\ref{fig: flowchart}) that integrates a novel Score-Based Line Fitting (SBLF) algorithm with geometric constraints informed by the known ramp model. More specifically, we leverage SBLF to precisely detect separators between the center ramp and the neighboring flares. These separators are then used to reliably identify the four corner points defining the center ramp by using their intersections with the ramp contour given by its convex hull (P1-P4 in Fig.~\ref{fig: ref_point}). The remaining two flare corners (P5-P6 in Fig.~\ref{fig: ref_point}) are identified based on the maximum projection distance of flare points to the separator lines as shown in Fig.~\ref{fig: ref_point}. Together, these six reference points (P1-P6 in Fig.~\ref{fig: ref_point}) minimally specify the reference points as they pertain to measurements. The following describes the SBLF algorithm and the corner detection process in detail.

\paragraph{Score-Based Line Fitting (SBLF)}\label{linefitting}
The Score-Based Line Fitting (SBLF) algorithm is specifically designed to identify clear geometric boundaries between point cloud segments, while maintaining robustness to outliers and local sparsity. Traditional approaches such as RANSAC~\cite{fischler1981random} and Support Vector Machines (SVM)~\cite{suthaharan2016support} often fall short in this context. RANSAC fits models by maximizing inlier consensus but requires careful tuning of distance thresholds that are highly sensitive to point cloud density, and also depends on tuning hyperparameters that control the softness of margin violations.

We designed SBLF to avoid such dependencies by directly focusing on the main objective of ramp component separation. This is done by exhaustively evaluating candidate boundary lines formed using point pairs, and assigning a score to the resulting separation. The score is computed as the number of points from each class that lie on their expected side of the line. For instance, when separating the center ramp from the left flare, SBLF identifies the line that maximizes left flare points on the left-hand side and center ramp points on the right-hand side of the separator. This simple yet effective strategy produces geometrically intuitive and visually clean boundaries without reliance on hyperparameters. Crucially, because the scoring is based on point distribution rather than distance metrics or learned decision boundaries, SBLF requires no hyperparameter tuning and remains robust across a wide range of point cloud local densities and noise conditions.

SBLF is formulated as follows. Given two point sets $A$ and $B$ to be separated, and a candidate line point set $C$, $\text{SBLF}(A, B, C)$ computes a separator line by evaluating all possible lines formed by pairs of points from $C$. For each candidate line $l: ax + by + c = 0$, we calculate a separation score:
\begin{align*}
    S(l) &= S_A + S_B, \\
    S_A &= |\{p \in A : ap_x + bp_y + c < 0\}|, \\
    S_B &= |\{p \in B : ap_x + bp_y + c > 0\}|,
\end{align*}
where $|\cdot|$ denotes set cardinality. The algorithm returns the line coefficients $(a, b, c)$ that maximize this score.

\paragraph{Component Separation}\label{linefitting2}
As part of the reference point detection process, we use SBLF to identify the primary geometric boundaries between the center ramp and the two flares. As discussed in the previous section, we also need the boundary between the center ramp bottom and unassigned points, for which we also leverage SBLF. To improve robustness, we find two separation lines for each boundary based on two different candidate line point sets, $C$, and then take an average of the two lines. To find the boundary between a flare $A$ and the center ramp $B$, we compute two candidate separator lines: one using $C=A$, and another with $C=B$. The final separator line is obtained by averaging these two lines:

\begin{align} \label{Eq:SepLineAve}
    l_{f} &= \text{SBLF}(P_{flare}, P_{middle}, P_{flare}), \notag\\
    l_{m} &= \text{SBLF}(P_{flare}, P_{middle}, P_{middle}), \notag\\
    l &= \frac{l_f + l_m}{2}.
\end{align}

To identify the bottom line of the center ramp using SBLF, we apply SBLF on two groups of points: center ramp points, and unlabeled ground points that lie between the left and right separator lines. We then further restrict to the bottom half‐space in the 3D elevation coordinate \(z\) (i.e., we only keep points whose height is below the centroid height of the center ramp) to further reduce the search space for SBLF and obtain $A$ and $B$. Like before, we average the two resulting SBLF lines when using $C=A$ and $C=B$ to finalize the bottom line fit.

\paragraph{Corner Point Detection}
Using the final separator lines (Eq. \ref{Eq:SepLineAve}), we identify six reference points (P1-P6 in Fig.~\ref{fig: ref_point}) that fully define the ramp geometric model. Specifically, four center ramp points, P1 to P4 as shown in Fig.~\ref{fig: ref_point},  are computed as intersections between the separator lines and the entire ramp's convex hull. The remaining two points on the flares, P5 and P6, are identified by maximum projection distance to the separators as shown in Fig.~\ref{fig: ref_point}:
\begin{equation*}
    p_{flare} = \text{argmax}_{p \in P_{flare}} \frac{|ap_x + bp_y + c|}{\sqrt{a^2 + b^2}}.
\end{equation*}

Furthermore, we establish a process that guarantees the assigned corner points P1--P6 are always consistently ordered regardless of the ramp's orientation. The algorithm works in two steps.

First, we determine the left--right orientation by leveraging the geometric relationship between separators and components. The separator lines between components are oriented such that when looking along each separator line normal vector, the respective flare component is always positioned on one predetermined side: left flare on the negative side of its separator, right flare on the positive side of its separator. This consistent orientation of separators provides an unambiguous definition of ``left'' versus ``right'' for the intersection points. We use the mean of the projections of all the points in the extracted ramp point cloud as the origin in a local coordinate system. Let $(a_L,b_L,c_L)$ and $(a_R,b_R,c_R)$ be the coefficients of the left and right separator lines, respectively.  We enforce that the left separator's unit normal vector $\mathbf{n}_L = \frac{(a_L,b_L)}{\sqrt{a_L^2 + b_L^2}}$ to point towards the center ramp and away from the left flare; if the sign of $\mathbf{n}_L  \cdot \bar{p}_\text{left}$ is positive—where $\bar{p}_\text{left}$ is the centroid of the left flare points in the coordinate system—then the centroid lies on the positive side of the separator line defined by $\mathbf{n}_L$, so we flip the separator coefficients: $(a_L,b_L,c_L) \to (-a_L,-b_L,-c_L)$ to ensure that the left flare remains on the negative side of its separator. We apply the same process to $\mathbf{n}_R = \frac{(a_R,b_R)}{\sqrt{a_R^2 + b_R^2}}$ to ensure the right flare stays on the positive side of its separator $(a_R,b_R,c_R)$.

Second, to identify ``top'' versus ``bottom'' corners, we rotate the two unit normal vectors $\mathbf{n}_L$ and $\mathbf{n}_R$ by $90^\circ$ counterclockwise to obtain vectors pointing towards the ramp's top edge from the bottom edge:
\[
\mathbf{u}_L = \frac{(-b_L, a_L)}{\sqrt{a_L^2+b_L^2}}, 
\quad
\mathbf{u}_R = \frac{(-b_R, a_R)}{\sqrt{a_R^2+b_R^2}}.
\]
Summing and normalizing these gives the unit ramp axis:
\[
\mathbf{v}
= \mathbf{u}_L + \mathbf{u}_R,
\qquad
\hat{\mathbf{r}}
= \frac{\mathbf{v}}{\|\mathbf{v}\|}.
\]
Vector $\hat{\mathbf{r}}$ consistently points from the ramp's bottom toward its top edge, regardless of global orientation. We then compute the centroid $\bar{p}$ of all intersection points and project each vector going from $\bar{p}$ to $p$ onto $\hat{\mathbf{r}}$:
\[
s = (p - \bar{p}) \cdot \hat{\mathbf{r}}.
\]

Points with larger $s$ values are classified as ``top'' corners because they are farther along the positive ramp axis, while points with smaller $s$ values are classified as ``bottom'' corners. This step is invariant to the ramp's absolute orientation in the coordinate system. This approach results in a consistent ordering of P1 through P6.

Finally, we map the 2D reference points back to 3D space. It is noted that the corner points are calculated and hence, their correspondence in the 3D space should be obtained via the nearest neighbor search in the original point cloud.

\subsection{Quality Assurance} \label{subsec:QC}
To improve the reliability of our measurements and prevent propagation of errors in the pipeline, we first remove ramps with insufficient point cloud density. Specifically, any ramp with an average point density below a predefined threshold is excluded to avoid inaccuracies in slope or length estimation due to sample sparsity or incomplete surface geometry. We apply density-based filtering at this stage because for a more relevant and accurate density estimation, we need to restrict the estimate to the individual ramp asset and not an entire point cloud. Further, density in terms of points per area is a better and more robust estimate compared to volumetric density, and as such, density-based filtering is applied once we geometrically refine the components.

The surface density estimate is obtained by first projecting the points for each component on the corresponding plane-fit, then obtaining the area of the convex hull of the component projections. Density can then be estimated by dividing the total number of considered points by the sum of the areas. While we can perform this for the final components, we leverage the initial OCSVM-identified coresets for the 2D density estimation instead in order to avoid potential error propagation, considering the coresets still incorporate many points and a substantial surface area sufficient for obtaining a robust estimate.

Additionally, to ensure geometric consistency in corner detection, we apply two filtering methods. First, we analyze the distribution of angles at the identified corners and exclude ramps where an angle deviates more than three standard deviations from the dataset mean. This works as a statistical outlier detection step where ramps with atypical corners are removed. Second, we enforce a parallelism constraint on the top and bottom edges of the center ramp. This is informed by our a priori knowledge about the construct of ADA ramps and similar constraints can be adapted to other assets. This constraint is critical because later these edges are used to approximate the landing and gutter regions to be used in their corresponding measurements.

\begin{figure*}[t!]
\centering
\includegraphics[width=\linewidth]{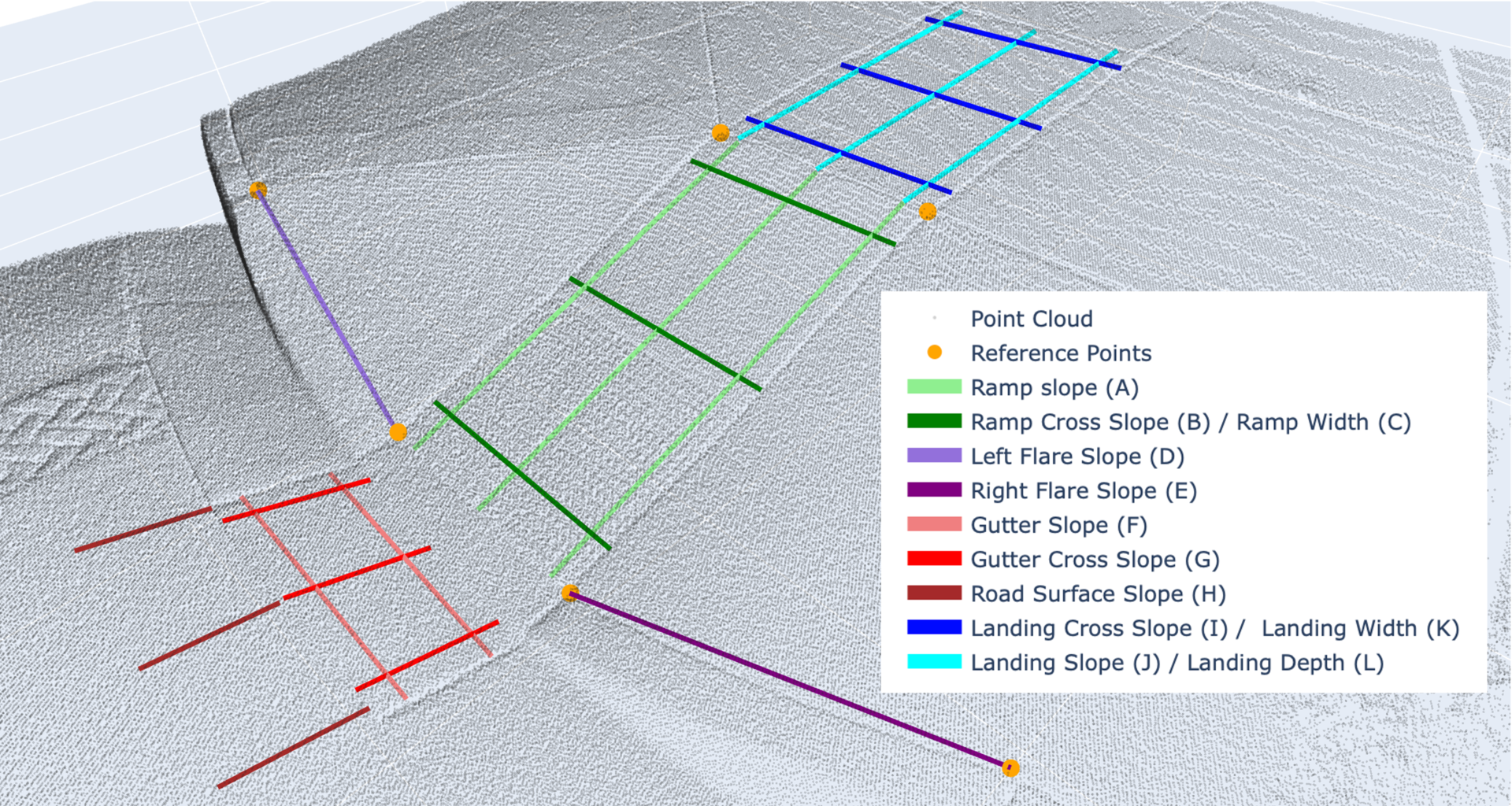}
\caption{Reference points and reference lines on a sample ramp point cloud. Orange markers indicate the reference points used for measurement extraction. Colored lines represent the initial reference lines where slopes and widths are calculated.}
\label{fig: ramp_reference}
\end{figure*}

\subsection{Measurement Extraction}
Using the extracted reference points (in this case corners), slope and width measurements can be computed following a process that mirrors Caltrans’ established guidelines. Adhering to these guidelines is important, as any modifications to the measurement procedure typically requires approval from relevant authorities, including legal and regulatory departments.

\subsubsection{Center Ramp Measurements}
As seen in Fig.~\ref{fig: ramp_reference}, the center ramp consists of four corner points. The objective is to measure ramp slope (A), ramp cross slope (B), and ramp width (C). 

Generate Reference Lines: Each edge of the center ramp is divided into 10 equal segments. To create initial reference lines, points at $1/10$, $5/10$, and $9/10$ of each line are connected with their corresponding points on the opposite side (green lines in Fig.~\ref{fig: ramp_reference}). 

Reference Line Refinement and Slope and Width Measurement: For each reference line, we collect 300 nearest points from the point cloud. Using these points, we fit a new line to ensure that all the relevant parts of the point cloud contribute to the measurements. Once the line is fitted, points that are more than $1/4$ inch away from the line are filtered out, adhering to Caltrans' measurement guidelines. This process of line fitting and point discarding is repeated iteratively until no points are removed. The final fitted line is used for slope and width measurements.

\subsubsection{Flare Measurements}
The two flares each consist of three corner points (Fig. \ref{fig: ref_point}). The objective is to measure the running slopes (indicated as Left Flare Slope and Right Flare Slope in Fig.~\ref{fig: ramp_reference}). For this purpose, we generate a reference line from the two bottom corner points (two purple lines in Fig.~\ref{fig: ramp_reference}). Using these points, we apply the same iterative process outlined for the center ramp measurement to extract slopes.

\begin{figure}[!t]
\centering
\includegraphics[width=\linewidth]{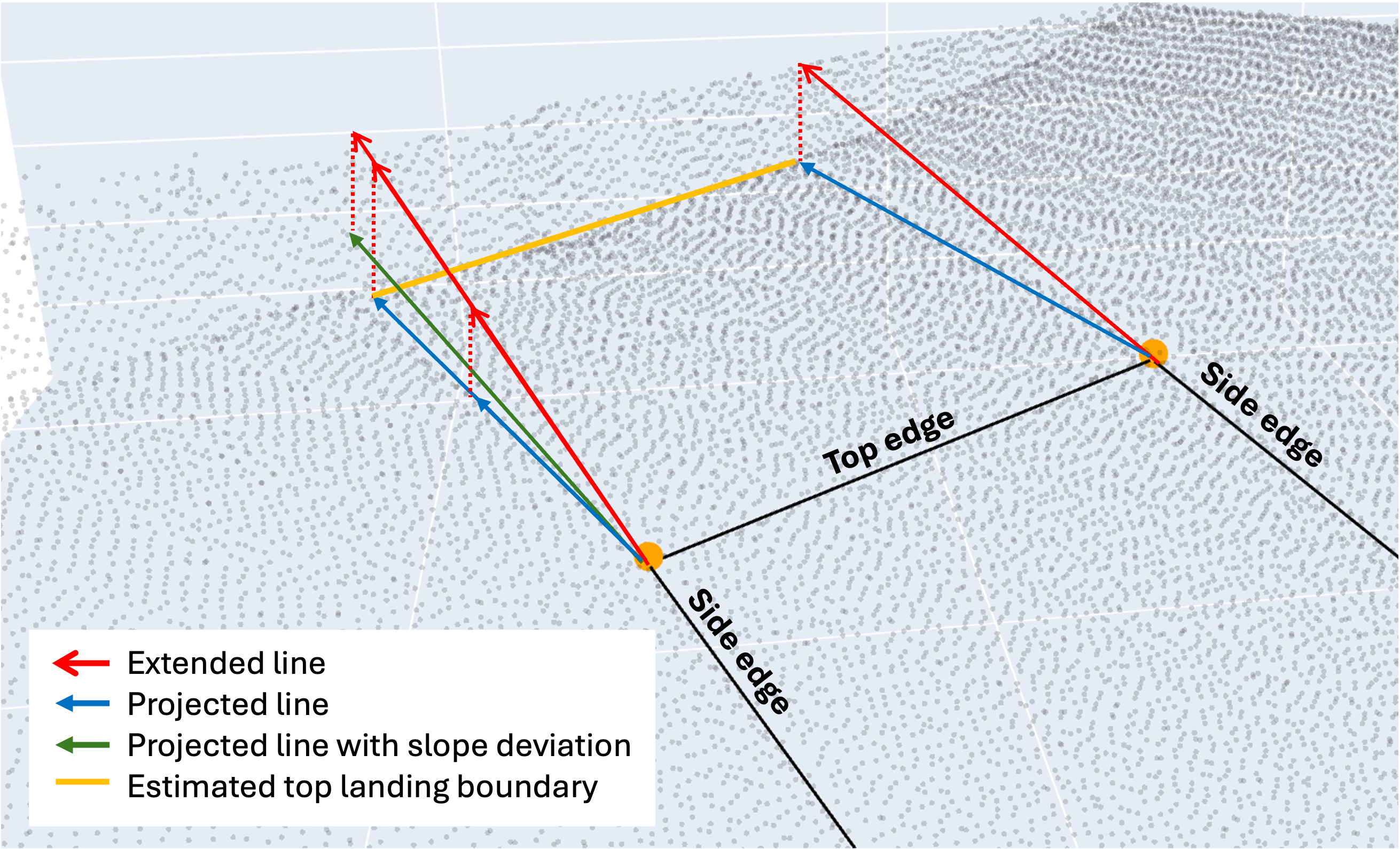}
\caption{Landing approximation from center ramp edges. The red arrows indicate the extended lines from the side edges of the center ramp, while the blue arrows show their vertical projections onto the point cloud surface. The green arrow represents a projected line with a detected slope deviation. The algorithm extends each side edge outward (red arrows) and monitors the slope of its projection (blue arrows). The algorithm terminates the extension when a sharp slope deviation is detected (i.e., rapid change in slope going from the blue arrow to the green arrow). The orange line marks the estimated top landing boundary connecting the two termination points.}
\label{fig: landing_approximation}
\end{figure}

\subsubsection{Landing and Gutter Measurement}
In our analysis pipeline, the landing and the gutter areas are approximated by imposing their minimal standard geometric definition directly onto the point cloud. For the landing area, we estimate the top landing boundary by extending the side edge of the center ramp outward and searching perpendicularly from the top edge (red arrows in Fig.~\ref{fig: landing_approximation}). This search is performed iteratively, extending the edge by a small fixed length at each step. At each iteration, the endpoint of the extended line is projected vertically (in the z-direction) onto the point cloud, forming a sloped line between the projected point and the top corner of the center ramp (shown as blue arrows in Fig.~\ref{fig: landing_approximation}). As the search progresses, the slopes of these consecutive lines are continuously monitored. When the difference between two consecutive slope calculations is larger than a pre-defined threshold, the algorithm marks the last valid projected point before the sharp difference as the top corner point of the landing. This procedure is executed on both sides of the center ramp to determine the full top landing boundary (orange line in Fig.~\ref{fig: landing_approximation}). Similarly, the gutter area is approximated using the bottom edge of the center ramp as reference. Using the boundary points derived from the approximation, we can calculate the slopes and lengths following the same measurement steps used for other ramp components. 

\section{Experiments}\label{sec: experiments}

\subsection{Dataset and Implementation Details}

\subsubsection{Detection}
The ramp detection dataset consists of 1604 2D images containing 3217 bounding box annotations of pedestrian crossing ramps at road intersections. Each image represents a crop of a street-level point cloud, projected top-down and converted into grayscale using normalized LiDAR intensity values. Our DeTR model is trained on an 8:2 split dataset with a batch size of 32 for 500 epochs. The Adam optimizer was used with a learning rate of $10^{-5}$ and a weight decay of $10^{-4}$.

\subsubsection{Segmentation}
The street-level crops of point cloud data, each containing multiple road intersections, were manually annotated, where various segments of each ramp were delineated. Through manual annotation each unlabeled point is assigned to one of six classes: center ramp, warning surface, left/right flares, landing, and gutter. The point cloud data associated with individual ramps are then extracted from the annotated data and, following 2D projection, are converted to $512\times 512$ grayscale images---we will later resize this to $256\times 256$ when feeding to the segmentation model, but the choice of $512\times 512$ resolution here is to allow for additional flexibility in exploring image segmentation protocols. To apply adaptive dilation, we divide each image into $32\times 32$ patches and consider $\kappa_{max}=50$ (Eq.~\ref{eq: kappa}).

The input images and masks for SAM are resized to $256\times 256$. Masks are converted to binary tensors with a channel dimension, where different channels represent different classes---which means that for each channel, we filter the mask image to represent the pixels matching the channel index, or equivalently class index, as 1, and the rest as 0). The class prediction for each channel is generated by its associated mask decoder. The model is initialized with pretrained weights. The image and prompt encoder weights are frozen during training and only the decoder modules are fine-tuned. The training process employs a hybrid loss consisting of Dice loss and cross-entropy loss with equal weights between the sigmoid-activated decoder outputs and targets. For an image of dimensions $H\times W$ with a binary $C$-channel target, the total loss can be written as
\begin{align*}
\mathcal{L}_{seg} &= \frac{1}{C} \sum_{c=1}^{C} \left( 1 - \frac{2 \sum_{i=1}^{H} \sum_{j=1}^{W} \hat{p}_{c,i,j} \, y_{c,i,j}}{\sum_{i=1}^{H} \sum_{j=1}^{W} (\hat{p}_{c,i,j}^2 + y_{c,i,j}) + \epsilon} \right) \notag\\
&\quad + \frac{1}{C \times H \times W} \sum_{c=1}^{C} \sum_{i=1}^{H} \sum_{j=1}^{W} CE_{c,i,j} ,
\label{eq:dicece}
\end{align*}
where $y$ and $\hat{y}$ respectively denote the target and the predicted values for a given pixel in a channel, $CE$ is the pixelwise cross-entropy loss, and $\epsilon=10^{-5}$ is a small constant for numerical stability. The pixel predictions are squared in the denominator of the Dice loss to stabilize the gradients and improve convergence, especially early on during training, where predictions are soft and uncertain (hence this is sometimes referred to as squared Dice loss)~\cite{vnet}. The model with a ViT Base~\cite{dosovitskiy2021an} image encoder is trained on 1541 samples with a batch size of 16 for 100 epochs. Adam optimizer is used with a learning rate of $10^{-5}$ and a weight decay of $10^{-4}$.

\subsubsection{Ramp Decomposition into Constituting Components}
After retrieving point cloud class assignments, we set the OCSVM slack variable $\nu=0.7$ for extracting internal component points for each of the three components (left/right flares, combined center ramp/warning surface). We find this to be a conservative value that consistently yields high-quality candidate points. We choose the RBF kernel coefficient $\gamma$ based on the default heuristic where $\gamma=1/{3\sigma^2_X}$ in which $\sigma^2_X$ is the variance of the (centered) input point cloud coordinates and $3$ in the denominator represents the space dimensionality. $\sigma^2_X$ acts as a natural scale measure, and dimensionality factor serves to counteract the growth of Euclidean distances with dimensionality to stabilize the kernel coefficient.

We set a threshold of $t=0.05\,ft$ (or $0.6\,in$) for the point-to-plane distance used for component assignment. We also set the neighborhood ball radius $r=3\,in$ for local plane-fitting, and the absolute cosine similarity threshold $s_{crit}=0.999$. We perform iForest with 100 trees and a contamination of $0.02$~\cite{iforest}. Finally, we set the DBSCAN distance parameter $\varepsilon =2\,ft$~\cite{dbscan}.

\subsection{Results}
The ramp detection results are presented in Table~\ref{tab:detection}. The two DeTR models demonstrated superior performance compared to the Faster R-CNN model. Specifically, the DeTR model with the ResNet-50 backbone achieved an mAP@0.50 of 0.866 and a Recall@$\text{IoU}\geq0.50$ of 0.765, while the ResNet-101 backbone further improved these metrics to an mAP@0.50 of 0.873 and a Recall@$\text{IoU}\geq0.50$ of 0.771. 

The use of overlapping crops during inference enhances overall detection reliability because the same ramp is seen multiple times under varying contexts or backgrounds. This strategy ensures that when a ramp is not detected in one crop, it can still be identified in other crops, reducing the chances of missed ramps.

\begin{table}
    \centering
    \caption{Detection Performance on test dataset.}
    \begin{tabular}{lcc}
        \toprule
        Model & mAP@0.50 & Recall@IoU$\geq$0.50 \\
        \midrule
        Faster R-CNN (ResNet-50) & 0.538 & -\\
        DeTR (ResNet-50) & 0.866 & 0.765 \\
        \textbf{DeTR (ResNet-101)} & \textbf{0.873} & \textbf{0.771} \\
        \bottomrule
    \end{tabular}
    \label{tab:detection}
\end{table}

The trained SAM model was tested on 61 ADA ramp assets. The average Dice score across the six asset regions (excluding the background class) was calculated as 0.857, with the highest score attributed to the flare segment (0.890) and the lowest score corresponding to the gutter (0.822). Detailed results are provided in Table~\ref{tab:segmentation}.
\begin{table}
    \centering
    \caption{Segmentation performance across asset regions.}
    \begin{tabular}{lclc}
        \toprule
        Segment & Dice Score & Segment & Dice Score \\
        \midrule
        Center Ramp & 0.842 & Warning Surface & 0.882 \\
        Right Flare & 0.880 & Left Flare & 0.890 \\
        Landing & 0.822 & Gutter & 0.822 \\
        \midrule
        \multicolumn{3}{l}{Mean (Excluding Background)} & 0.857 \\
        \bottomrule
    \end{tabular}
    \label{tab:segmentation}
\end{table}

For quality control, density-based filtering is applied to 46 ramps whose components are geometrically refined in the previous step, disqualifying 11 ramps due to insufficient point cloud resolution, leaving 35 ramps for further analysis. Among these, another 4 were disqualified per angle-based statistical filter using a 3-standard-deviation threshold, and 11 were removed by the parallelism constraint on opposing ramp edges. In total, 20 ramps passed both stages of quality control and were retained for the final evaluation. Further discussion of quality control rationale, trade-offs, and implications is provided in Section \ref{sec: discussion}.

\begin{figure}[!t]
\centering
\includegraphics[width=\linewidth]{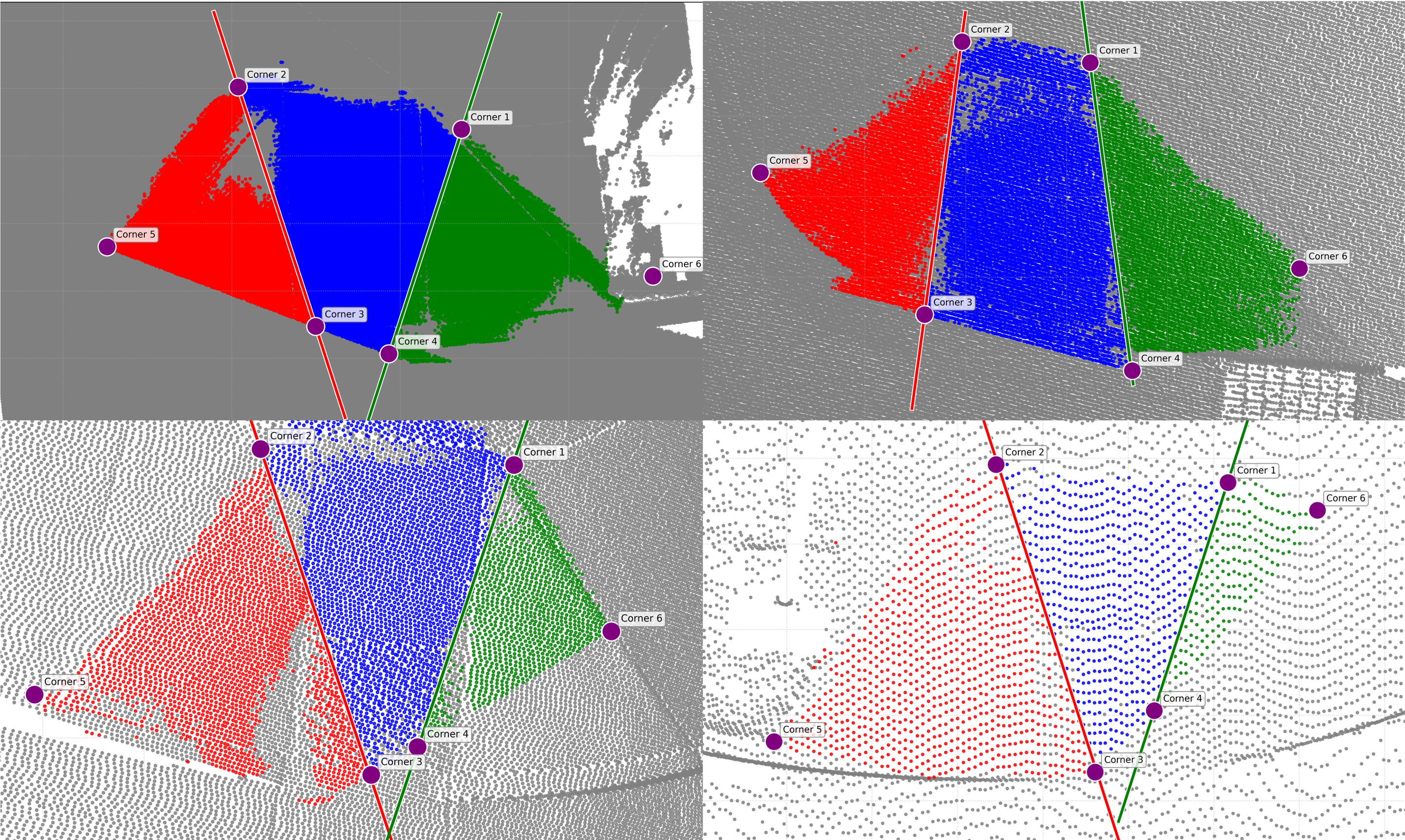}
\caption{Examples of excluded outlier ramps. These ramps exhibit irregular segmentation or corner placements that deviate significantly from expected geometric patterns.}
\label{fig: outlier_example}
\end{figure}

Figure~\ref{fig: outlier_example} illustrates an example of excluded disqualified ramps, which shows irregular segmentation or corner placements, leading to unexpected geometry such as extreme angles as discussed in Section \ref{subsec:QC}.  

\begin{table}
    \centering
    \caption{Measurement features and compliance standards.}
    \begin{tabular}{lcc}
        \toprule
        Feature & Compliance standard\\
        \midrule
        A - Ramp Slope (\%) & $\leq7.7\%$ \\
        B - Ramp Cross Slope (\%) & $\leq1.7\%$ \\
        C - Ramp Width (inches) & $\geq49.75\text{"}$ \\
        D - Left Flare Slope (\%) & $\leq9.2\%$ \\
        E - Right Flare Slope (\%) & $\leq9.2\%$ \\
        \midrule
        F - Gutter Slope (\%) & $\leq1.7\%$ \\
        G - Gutter Cross Slope (\%) & $\leq5.2\%$ \\
        H - Road Surface Cross Slope (\%) & $\leq5.2\%$ \\
        \midrule
        I - Top Landing Cross Slope (\%) & $\leq1.7\%$ \\
        J - Top Landing Slope (\%) & $\leq1.7\%$ \\
        K - Top Landing Width (inches) & $\geq49.75\text{"}$ \\
        L - Top Landing Depth (inches) & $\geq49.75\text{"}$ \\
        \bottomrule
    \end{tabular}
    \label{tab: labels-compliance-standards}
\end{table}

To evaluate our automated measurement pipeline, we compared the ADA compliance assessment results against manual measurements (see Appendix for manual field measurement details). We conducted field measurements on 16 ramps in the city of Woodland, CA, selected from the subset that passed quality control and were accessible based on geographic proximity and ease of access for our team. The comparison was performed across twelve primary ADA ramp features, each of which is used by Caltrans in compliance assessment as illustrated in Table \ref{tab: labels-compliance-standards}. Each feature comprises one or more measurements (e.g., A1–A3, B1–B3, etc.), resulting in a total of 31 sub-feature measurements used for detailed evaluation. For each feature, we calculated the average measurement difference and compliance consistency between the automated and manual assessments. 

\begin{figure*}[t!]
\centering
\includegraphics[width=\linewidth]{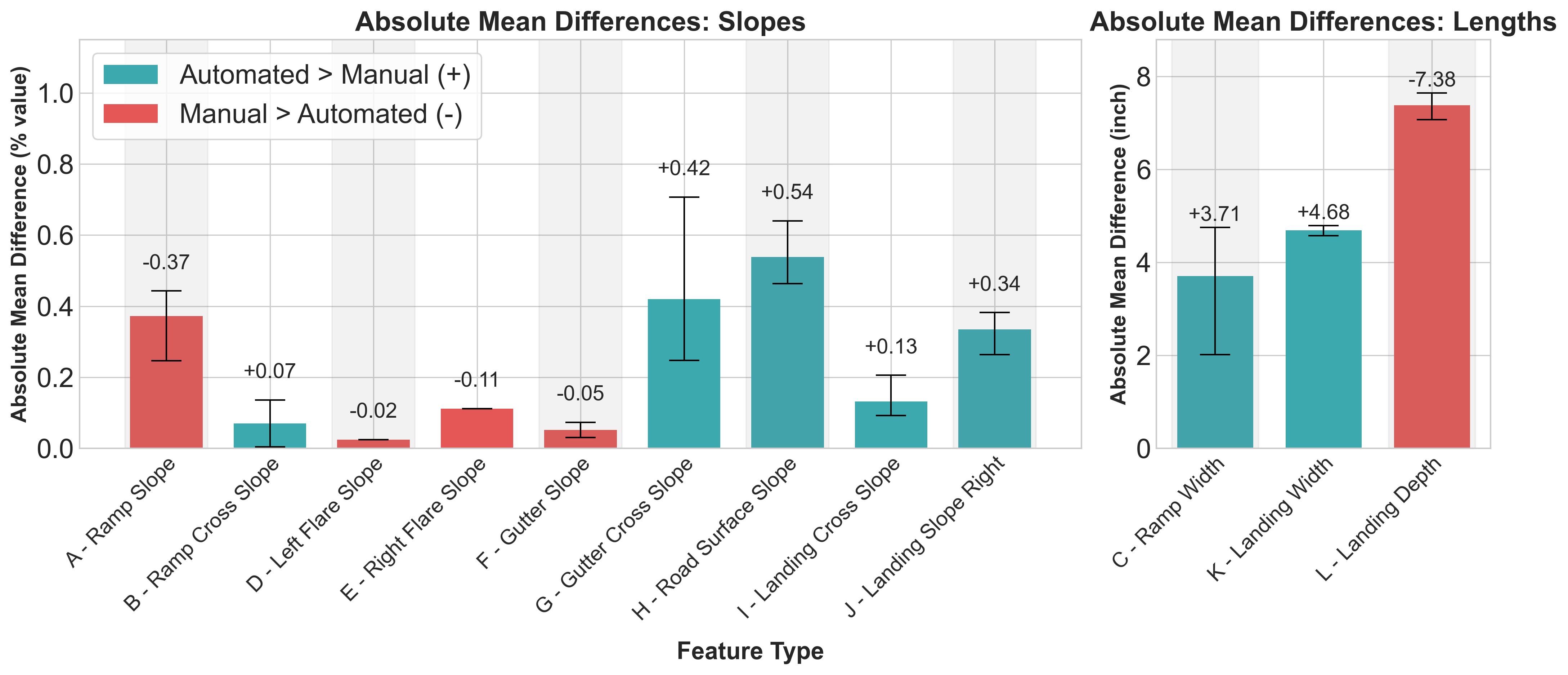}
\caption{Absolute mean differences between automated and manual measurements across ADA ramp features. Differences are separated into slope-based (\% value) and length-based (inches) measurements. Positive values (blue) indicate overestimation by the automated pipeline, while negative values (red) indicate underestimation. Black ranges represent the minimum and maximum observed differences for each feature.}
\label{fig: measurement-results-diff}
\end{figure*}

Figure \ref{fig: measurement-results-diff} illustrates the absolute mean differences between our automated measurements and manual measurements for each ADA ramp feature. For slope-based features (left), the differences remain relatively small, generally under 1\%, indicating strong alignment between automated and manual processes. The largest slope deviation was observed for the road surface slope (feature H, Fig. \ref{fig: ramp_reference}), at approximately +0.54\%. For length-based features (right), larger discrepancies were found, particularly in landing depth (feature L, Fig. \ref{fig: ramp_reference}), where manual annotations tend to exceed automated estimates by an average of 7.88 inches. This is due to the approximation involved in estimating the top landing line, where small artifacts or irregularities in the surface geometry may lead to inconsistent depth calculations between manual and automated methods.

\begin{figure}[h!]
\centering
\includegraphics[width=\linewidth]{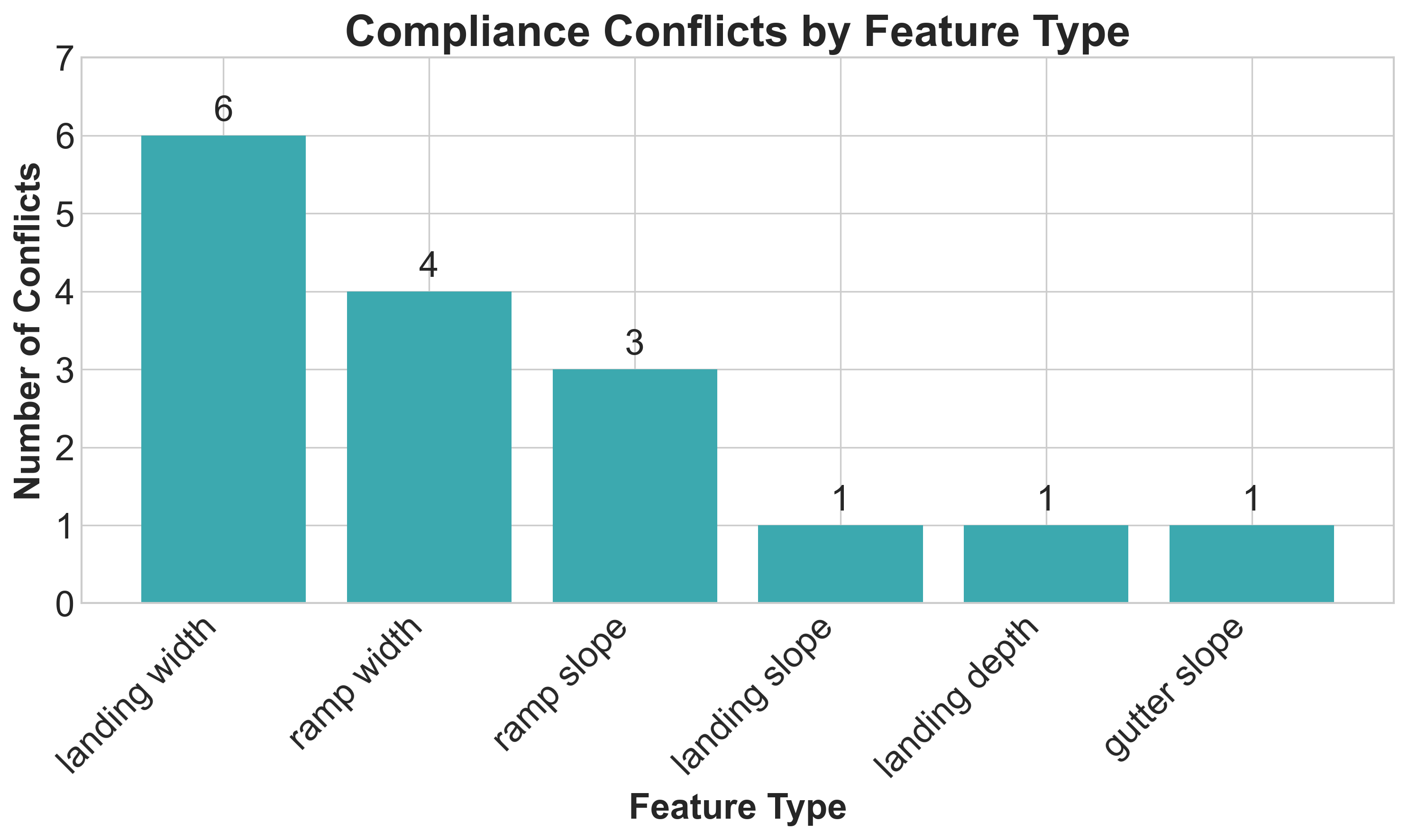}
\caption{Analysis of compliance consistency between automated and manual ADA ramp assessments. The bar chart shows the number of compliance conflicts per feature type, with the most frequent discrepancies observed in landing width and center ramp width.}
\label{fig: measurement-results-compliance}
\end{figure}

Since the final decision on ADA compliance is binary, we further evaluated the practical utility of our system by comparing binary compliance assessments from the automated pipeline with those obtained through manual evaluation. Figure \ref{fig: measurement-results-compliance} highlights the number of compliance conflicts per feature type. The majority of disagreements occurred in width-based features, particularly landing width (6 cases) and center ramp width (4 cases), followed by center ramp slope (3 cases). These features are often sensitive to boundary definitions, and the discrepancies can arise from local surface irregularities not captured uniformly in manual inspections especially when concrete lines or transitions are ambiguous. Such inconsistencies underscore the benefit of a consistent, geometry-based approach, as used in our automated pipeline. It should be noted that in cases where automated assessment flags a ramp as non-compliant, manual follow-up can be limited to the specific feature(s) that failed quality checks or compliance thresholds. This targeted verification further reduces the overall manual effort compared to traditional full-ramp assessments.

\begin{figure*}[h!]
\centering
\includegraphics[width=\linewidth]{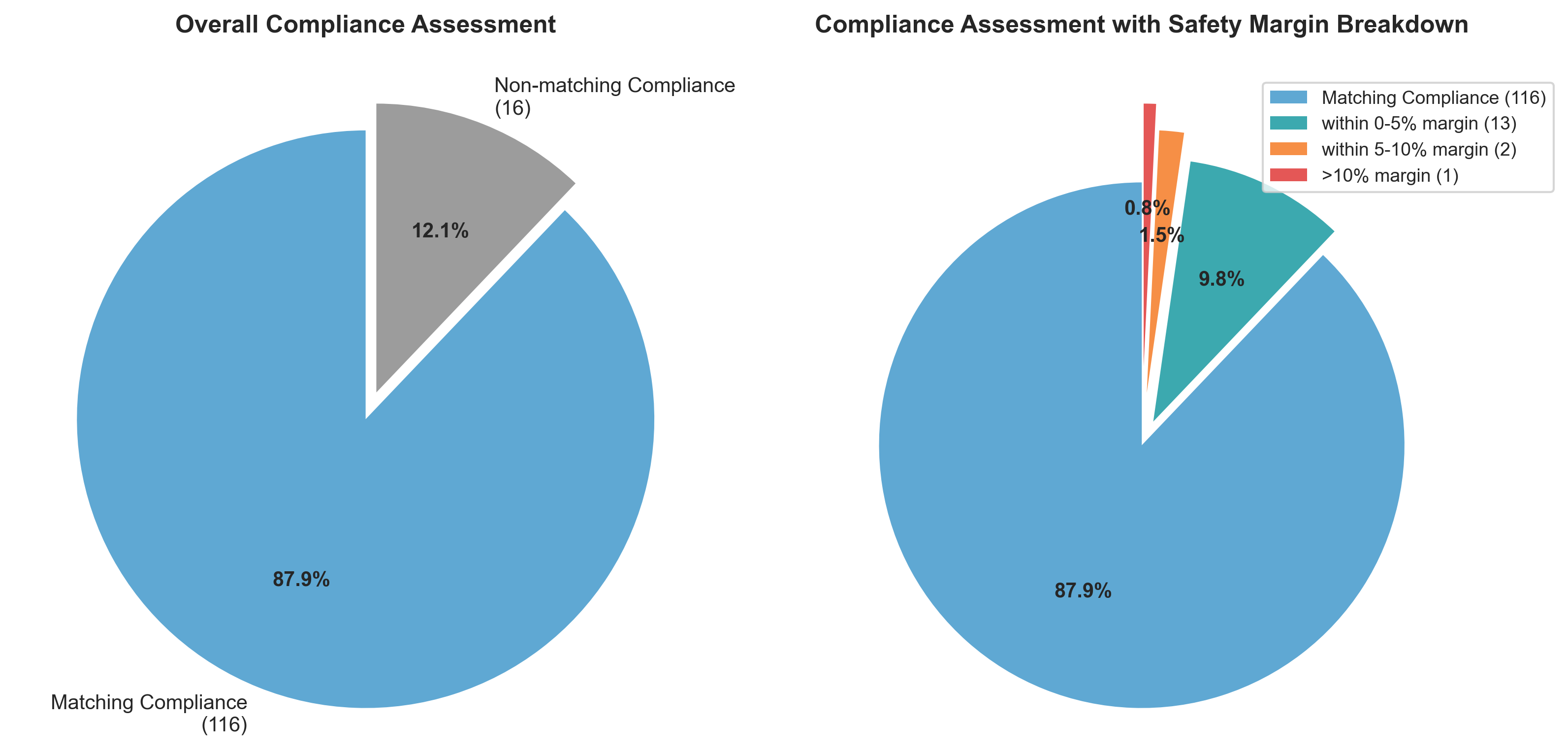}
\caption{Left: overall compliance agreement between automated and manual assessments without any tolerance margin applied.
Right: breakdown of the 12.1\% non-matching compliance results after introducing progressive tolerance margins of 5\%, 10\%, and beyond. Most conflicts fell within a 5\% margin, suggesting that minor deviations in manual measurement could explain a majority of the disagreements.}
\label{fig: measurement-results-compliance-breakdown}
\end{figure*}

To account for human error in manual measurements or inaccuracies due to inconsistent point clouds, we define a tolerance margin around the compliance threshold \( T \) to assess whether the disagreement is significant. Let the margin be defined as:
\[
\delta = T \times \frac{p}{100},
\]
where \( p \) is our defined margin percentage (e.g., \( p = 5 \)). We define the acceptable range as:
\[
[T - \delta,\, T + \delta].
\]
If both the automated measurement \( A \) and the manual measurement \( M \) fall within this range, i.e.:
\[
T - \delta \leq A \leq T + \delta \quad \text{and} \quad T - \delta \leq M \leq T + \delta,
\]
then the disagreement is considered within tolerance and not treated as a conflict.
We set tolerance margins of 5\% and 10\% to evaluate the robustness of compliance consistency. As shown in Fig. \ref{fig: measurement-results-compliance-breakdown}, the left chart shows that 87.9\% of feature assessments were consistent with manual measurements. Among the remaining 12.1\%, the breakdown (right) shows that most discrepancies (13 out of 16) fell within a 5\% tolerance margin, and only a single instance exceeded a 10\% margin. This analysis suggests that a majority of the conflicts stem from minor and likely acceptable human variation, rather than fundamental measurement errors in the automated system. The results demonstrate that our method remains robust and reliable, even under modest margin thresholds.

\section{Discussions and Future Work}\label{sec: discussion}

As noted earlier, over half of the detected ramps were disqualified through quality control. This reflects a deliberate trade-off in favor of measurement reliability, essential in high-stakes applications such as ADA compliance, where false negatives may carry legal and safety implications. Of the disqualified ramps, nearly half were excluded due to insufficient point cloud density, a limitation specific to re-used datasets not originally captured with automated assessment in mind. This highlights an opportunity: by establishing minimum data quality standards, future point cloud collections by Departments of Transportation can be better aligned with automation needs, significantly reducing disqualification rates. Even under current conditions, our framework successfully automates the analysis of nearly half of the ramps, representing a substantial reduction in manual labor and associated costs. It is also important to note that among disqualified ramps, those rejected due to severe geometric deviations were all identified as ADA non-compliant through manual inspection. This outcome is expected, as such quality control failures can serve as proxies for severe non-compliance, where significant structural irregularities either hinder reliable geometric analysis or produce results that clearly warrant immediate field inspection. Rather than reflecting a shortcoming, these disqualifications underscore the value of quality control in flagging severely problematic cases early in the process. 

While our automated pipeline demonstrates strong results, there are several limitations worth considering. First, the accuracy of the automated measurements is inherently dependent on the quality and density of the input point clouds. Sparse or noisy data, often caused by occlusions, or poor sensor calibration, can lead to misidentification of ramp boundaries or make slope/length measurements error-prone. Although we apply outlier filtering, this step cannot fully mitigate the effects of low-quality data. Second, several ramp components such as landing edges and the gutter transitions may lack explicit visual cues in the field. In those cases, both manual and automated methods rely on approximations. In particular, our method uses geometric projections and slope-based heuristics to infer feature boundaries, which may diverge from human interpretation in ambiguous regions where markings are faded or surface continuity is disrupted. Another limitation is that while our approximated landing and gutter boundaries work reasonably well, they rely on geometric regularity and slope transitions that may not generalize well to highly irregular ramp designs. In such cases, the automated system might either fail to detect the intended boundary or assign it incorrectly.

We further note that field measurements are not entirely geometric and in practice, ramp visual cues such as boundaries factor into the measurements. For instance, ramp cross slope direction is defined relative to the visible ramp boundaries. Consequently, there exists an inherent difference between the field measurements and those obtained by the proposed method when separation of ramp components per visual references and local surface topography do not align. As discussed earlier, we used visual segmentation to retrieve a relatively crude prior for geometry-based component decomposition downstream. Given the aforementioned inherent difference, if a closer match with field measurements is desired, it can be obtained by relying more closely on the visual segmentation output rather than geometric modeling. Of course as previously stated, an output based on visual segmentation would not align with actual geometric features as closely, which is the de facto motivation behind the measurements in the first place. 

Future research directions include extending the pipeline to handle partially occluded assets and further exploring multimodal data, for instance combining LiDAR with high-resolution images.

\section{Conclusion}\label{sec: conclusion}
This paper presents a framework for automated surveying and compliance assessment of infrastructure assets, demonstrated through the case study of curb ramps for ADA compliance. By integrating modern deep learning techniques for object detection and segmentation with rule-based protocols and mathematical modeling, the proposed pipeline detects ramps, identifies and refines their components, and extracts key reference points for reliable measurement. Experimental validation across real-world ramps and against field measurements highlights the effectiveness of our hybrid approach.

In addition to the methodological contributions, we are releasing an annotated dataset of over 1600 assets, along with an open-source code and manual point-cloud and field measurements for several ADA ramps. This enables other researchers and practitioners to build upon our approach or develop alternative techniques to advance the field. While the present work focuses on specific ADA ramp requirements, the underlying framework and ideas, i.e. uniting AI-powered detection and segmentation with mathematical modeling and rule-based systems, has broader applicability for automating surveying tasks. We anticipate that these findings will encourage broader adoption of digital data for automation within transportation agencies and beyond, reducing manual overhead, enhancing quality control, enabling scalable deployment across large areas, and ultimately contributing to safer and more accessible urban environments.

{\appendix[Manual Field Measurements]

All measurements were taken according to Caltrans training and the procedures outlined in Form DOT CEM‑5773ADE (Rev. 07/2024)~\cite{CEM5773ADE}.

\subsection{Instrumentation}

The below instruments were used for conducting field measurements:
\begin{itemize}
  \item 10‑ft (3-m) tape measure (additive inches);
  \item 2‑ft Stabila digital (smart) level;
  \item 4‑ft Stabila digital (smart) level.
\end{itemize}

Prior to starting measurements on each new ADA ramp, the smart levels were both reset and calibrated to ensure accuracy. The reference document provided by Caltrans requires that slopes be recorded as percent grade. Given the higher precision of Stabila smart levels when performing degree measurement (0.05°) compared to percent grade (1\%) for the angle regime under consideration, we chose to record the angles in degrees first, and later convert them to percent grade using the following equation:
\begin{align*}
  \text{percent grade}=\tan(\theta)\times 100,
\end{align*}
where $\theta$ is the recorded angle in degrees.

\subsection{Center Ramp and Gutters}

Slope readings for the center ramp and gutter slopes were taken by placing a smart level directly on the pavement surface in the locations specified by Form CEM‑5773ADE. Table~\ref{tab:orientation} specifies the orientation of the level relative to the curb for each slope measurement.

\begin{table}[h]
  \centering
  \caption{Level orientation relative to curb for slope measurements.}
  \begin{tabular}{ll}
    \toprule
    \textbf{Measurement} & \textbf{Level orientation} \\
    \midrule
    Ramp slope (A1--A3)                      & Perpendicular \\
    Ramp cross slope (B1--B3)                & Parallel      \\
    Flare slopes (D1, E1)                    & Parallel      \\
    Gutter slope (F1, F2)                    & Parallel      \\
    Gutter cross slope (G1--G3)              & Perpendicular \\
    Road surface cross slope (H1--H3)        & Perpendicular \\
    Landing cross slope (I1--I3)             & Parallel      \\
    Landing slope (J1--J3)                   & Perpendicular \\
    \bottomrule
  \end{tabular}
  \label{tab:orientation}
\end{table}

The 2-ft or 4-ft level was selected at the discretion of the operator based on the available surface area, which sometimes differed between ramps. Ramp width (C1--C3) was measured with the measuring tape at the same locations as the corresponding slope readings and recorded to the nearest \(\tfrac{1}{4}\) inch.

\subsection{Flares}
Flare slopes (D1, E1) were measured using the 4-ft level. One end of the level was placed approximately 1 inch away from the bottom corner of the flare nearest to the inside (body) of the center ramp. The other end was angled upwards toward the outermost corner. If the lower flare edge was straight, the level was placed directly along that edge. If, on the other hand, the edge was curved, the level was aligned parallel to the shortest straight path between the two corners.

\subsection{Landing}

The landing was measured with a methodology similar to that of the center ramp. The landing cross slope (I1--I3) was measured with the 4-ft level facing parallel to the curb, and the landing slope (J1--J3) was measured with the 4-ft level facing perpendicular to the curb. At the same spots used to measure slope, we also used a tape measure to record the landing width (K1--K3) and depth (L1--L3), noting each to the nearest \(\tfrac{1}{4}\) inch.}

\bibliographystyle{IEEEtran}  % or another style that supports authoryear
\bibliography{Ref}     % references.bib is your .bib file name

\begin{thebibliography}{10}
\providecommand{\url}[1]{#1}
\csname url@samestyle\endcsname
\providecommand{\newblock}{\relax}
\providecommand{\bibinfo}[2]{#2}
\providecommand{\BIBentrySTDinterwordspacing}{\spaceskip=0pt\relax}
\providecommand{\BIBentryALTinterwordstretchfactor}{4}
\providecommand{\BIBentryALTinterwordspacing}{\spaceskip=\fontdimen2\font plus
\BIBentryALTinterwordstretchfactor\fontdimen3\font minus \fontdimen4\font\relax}
\providecommand{\BIBforeignlanguage}[2]{{%
\expandafter\ifx\csname l@#1\endcsname\relax
\typeout{** WARNING: IEEEtran.bst: No hyphenation pattern has been}%
\typeout{** loaded for the language `#1'. Using the pattern for}%
\typeout{** the default language instead.}%
\else
\language=\csname l@#1\endcsname
\fi
#2}}
\providecommand{\BIBdecl}{\relax}
\BIBdecl

\bibitem{biondini2023life}
F.~Biondini, D.~M. Frangopol \emph{et~al.}, ``Life-cycle of structures and infrastructure systems,'' in \emph{Proceedings Of The Eighth International Symposium On Life-Cycle Civil Engineering}.\hskip 1em plus 0.5em minus 0.4em\relax CRC Press/Balkema, 2023.

\bibitem{doi:10.1061/(ASCE)0887-3801(2008)22:3(216)}
M.~R. Halfawy, ``Integration of municipal infrastructure asset management processes: Challenges and solutions,'' \emph{Journal of Computing in Civil Engineering}, vol.~22, no.~3, pp. 216--229, 2008.

\bibitem{Economicefficiency}
I.~Openko, O.~Shevchenko, Y.~Stepchuk, R.~Tykhenko, O.~Tykhenko, and A.~Horodnycha, ``Economic efficiency of methods for surveying rural infrastructure: assessment of accuracy, cost and duration,'' in \emph{23rd International Scientific Conference Engineering for Rural Development}, 05 2024.

\bibitem{jacobs2009ada}
\BIBentryALTinterwordspacing
J.~E. Group, ``Ada transition plans: A guide to best management practices,'' National Cooperative Highway Research Program (NCHRP), Transportation Research Board, Baltimore, MD, Tech. Rep., 2009, nCHRP Project Number 20-7 (232). [Online]. Available: \url{https://www.fhwa.dot.gov/indiv/docs/ada_transition_plans_report.pdf}
\BIBentrySTDinterwordspacing

\bibitem{fhwa2022nbi}
\BIBentryALTinterwordspacing
{Federal Highway Administration, Office of Bridges and Structures}, ``Specifications for the national bridge inventory,'' U.S. Department of Transportation, Tech. Rep. FHWA-HIF-22-017, March 2022, accessed: 2025-04-08. [Online]. Available: \url{https://www.fhwa.dot.gov/bridge/snbi/snbi_march_2022_publication.pdf}
\BIBentrySTDinterwordspacing

\bibitem{pierce2013practical}
\BIBentryALTinterwordspacing
L.~M. Pierce, G.~McGovern, and K.~A. Zimmerman, ``Practical guide for quality management of pavement condition data collection,'' U.S. Department of Transportation, Federal Highway Administration, Tech. Rep., 2013, accessed: 2025-05-04. [Online]. Available: \url{https://www.fhwa.dot.gov/pavement/management/qm/data_qm_guide.pdf}
\BIBentrySTDinterwordspacing

\bibitem{mallela2013mire}
\BIBentryALTinterwordspacing
J.~Mallela, S.~Sadasivam, R.~L. Becker, D.~Raghunathan, N.~Lefler, R.~Fiedler, and H.~W. McGee, ``Mire data collection guidebook,'' United States Federal Highway Administration, Office of Safety, Tech. Rep. FHWA-SA-13-009, June 2013, accessed: 2025-04-08. [Online]. Available: \url{https://rosap.ntl.bts.gov/view/dot/49490}
\BIBentrySTDinterwordspacing

\bibitem{torres2018automation}
\BIBentryALTinterwordspacing
H.~N. Torres, J.~M. Ruiz, G.~K. Chang, J.~L. Anderson, and S.~I. Garber, ``Automation in highway construction part i: Implementation challenges at state transportation departments and success stories,'' United States Federal Highway Administration, Office of Infrastructure Research and Development, Tech. Rep. FHWA-HRT-16-030, October 2018, accessed: 2025-04-08. [Online]. Available: \url{https://rosap.ntl.bts.gov/view/dot/41947}
\BIBentrySTDinterwordspacing

\bibitem{CEM5773ADE}
\BIBentryALTinterwordspacing
{California Department of Transportation}, ``Form dot cem-5773ade: Curb ramp (case a, d, or e) americans with disabilities act (ada) compliance inspection report,'' California Department of Transportation Forms, accessed: April 7, 2025. [Online]. Available: \url{https://forms.dot.ca.gov/v2Forms/servlet/FormRenderer?frmid=DOTCEM5773ADE}
\BIBentrySTDinterwordspacing

\bibitem{10641055}
J.~K. Liu, R.~Qin, and S.~Song, ``Automated deep learning-based point cloud classification on usgs 3dep lidar data using a transformer,'' in \emph{IGARSS 2024 - 2024 IEEE International Geoscience and Remote Sensing Symposium}, 2024, pp. 8518--8521.

\bibitem{rs12223757}
H.~Kim and C.~Kim, ``Deep-learning-based classification of point clouds for bridge inspection,'' \emph{Remote Sensing}, vol.~12, no.~22, 2020.

\bibitem{8099499}
R.~Q. Charles, H.~Su, M.~Kaichun, and L.~J. Guibas, ``Pointnet: Deep learning on point sets for 3d classification and segmentation,'' in \emph{2017 IEEE Conference on Computer Vision and Pattern Recognition (CVPR)}, 2017, pp. 77--85.

\bibitem{3326943.3327020}
Y.~Li, R.~Bu, M.~Sun, W.~Wu, X.~Di, and B.~Chen, ``Pointcnn: convolution on x-transformed points,'' in \emph{Proceedings of the 32nd International Conference on Neural Information Processing Systems}, ser. NIPS'18.\hskip 1em plus 0.5em minus 0.4em\relax Red Hook, NY, USA: Curran Associates Inc., 2018, p. 828–838.

\bibitem{10.1145/3326362}
Y.~Wang, Y.~Sun, Z.~Liu, S.~E. Sarma, M.~M. Bronstein, and J.~M. Solomon, ``Dynamic graph cnn for learning on point clouds,'' \emph{ACM Trans. Graph.}, vol.~38, no.~5, Oct. 2019.

\bibitem{Kumar17072020}
B.~L. Bhavesh~Kumar, Gaurav~Pandey and S.~C. Misra, ``A framework for automatic classification of mobile lidar data using multiple regions and 3d cnn architecture,'' \emph{International Journal of Remote Sensing}, vol.~41, no.~14, pp. 5588--5608, 2020.

\bibitem{suthaharan2016support}
S.~Suthaharan, ``Support vector machine,'' \emph{Machine learning models and algorithms for big data classification: thinking with examples for effective learning}, pp. 207--235, 2016.

\bibitem{10887345}
J.~Ye, X.~Liu, H.~Madhusudanan, Y.~Wang, J.~Zhu, Y.~Wang, C.~Ru, X.~Liu, and Y.~Sun, ``Automatic point cloud clustering for surface defect diagnosis,'' \emph{IEEE Transactions on Automation Science and Engineering}, vol.~22, pp. 12\,538--12\,547, 2025.

\bibitem{10988584}
Y.~Tian, Y.~F. Lin, and S.~Ma, ``Attention mechanism based pipe recognition network with a hybrid dataset,'' \emph{IEEE Transactions on Automation Science and Engineering}, pp. 1--1, 2025.

\bibitem{9259076}
H.~V. Dang, H.~Tran-Ngoc, T.~V. Nguyen, T.~Bui-Tien, G.~De~Roeck, and H.~X. Nguyen, ``Data-driven structural health monitoring using feature fusion and hybrid deep learning,'' \emph{IEEE Transactions on Automation Science and Engineering}, vol.~18, no.~4, pp. 2087--2103, 2021.

\bibitem{9374750}
T.~Bruggemann, ``Automated feature-driven flight planning for airborne inspection of large linear infrastructure assets,'' \emph{IEEE Transactions on Automation Science and Engineering}, vol.~19, no.~2, pp. 804--817, 2022.

\bibitem{qi2017pointnetdeephierarchicalfeature}
C.~R. Qi, L.~Yi, H.~Su, and L.~J. Guibas, ``Pointnet++: deep hierarchical feature learning on point sets in a metric space,'' in \emph{Proceedings of the 31st International Conference on Neural Information Processing Systems}, ser. NIPS'17.\hskip 1em plus 0.5em minus 0.4em\relax Red Hook, NY, USA: Curran Associates Inc., 2017, p. 5105–5114.

\bibitem{Zhang2023LiDAR}
\BIBentryALTinterwordspacing
Y.~Zhang, ``Deep learning with lidar point cloud data for automatic roadway health monitoring,'' The University of Texas at Austin; California State Polytechnic University, Pomona, Tech. Rep., May 2024, center for Understanding Future Travel Behavior and Demand. [Online]. Available: \url{https://tbd.ctr.utexas.edu/wp-content/uploads/2024/01/Exhibit-D_Y.Zhang_point_cloud_data.pdf}
\BIBentrySTDinterwordspacing

\bibitem{doi:10.1139/cjce-2024-0312}
H.~Jiang, H.~Elmasry, S.~Lim, and K.~El-Basyouny, ``Utilizing deep learning models and lidar data for automated semantic segmentation of infrastructure on multilane rural highways,'' \emph{Canadian Journal of Civil Engineering}, vol.~0, no.~0, p. null, 0.

\bibitem{https://doi.org/10.1111/j.1467-8667.2011.00727.x}
C.~Zhang and A.~Elaksher, ``An unmanned aerial vehicle-based imaging system for 3d measurement of unpaved road surface distresses,'' \emph{Computer-Aided Civil and Infrastructure Engineering}, vol.~27, no.~2, pp. 118--129, 2012.

\bibitem{SIEBERT20141}
S.~Siebert and J.~Teizer, ``Mobile 3d mapping for surveying earthwork projects using an unmanned aerial vehicle (uav) system,'' \emph{Automation in Construction}, vol.~41, pp. 1--14, 2014.

\bibitem{RAKHA2018252}
T.~Rakha and A.~Gorodetsky, ``Review of unmanned aerial system (uas) applications in the built environment: Towards automated building inspection procedures using drones,'' \emph{Automation in Construction}, vol.~93, pp. 252--264, 2018.

\bibitem{6917066}
P.~Prasanna, K.~J. Dana, N.~Gucunski, B.~B. Basily, H.~M. La, R.~S. Lim, and H.~Parvardeh, ``Automated crack detection on concrete bridges,'' \emph{IEEE Transactions on Automation Science and Engineering}, vol.~13, no.~2, pp. 591--599, 2016.

\bibitem{6705706}
R.~S. Lim, H.~M. La, and W.~Sheng, ``A robotic crack inspection and mapping system for bridge deck maintenance,'' \emph{IEEE Transactions on Automation Science and Engineering}, vol.~11, no.~2, pp. 367--378, 2014.

\bibitem{10855578}
J.~Feng, B.~Shang, E.~Hoxha, C.~Hernández, Y.~He, W.~Wang, and J.~Xiao, ``Robotic inspection and data analytics to localize and visualize the structural defects of concrete infrastructure,'' \emph{IEEE Transactions on Automation Science and Engineering}, pp. 1--1, 2025.

\bibitem{doi:10.1061/(ASCE)IS.1943-555X.0000353}
D.~Lattanzi and G.~Miller, ``Review of robotic infrastructure inspection systems,'' \emph{Journal of Infrastructure Systems}, vol.~23, no.~3, p. 04017004, 2017.

\bibitem{kumar2025designimplementationdualuncrewed}
D.~Kumar, A.~Ghorbanpour, K.~Yen, and I.~Soltani, ``Design and implementation of a dual uncrewed surface vessel platform for bathymetry research under high-flow conditions,'' 2025, arXiv:2502.12539 [cs.RO].

\bibitem{carion2020end}
N.~Carion, F.~Massa, G.~Synnaeve, N.~Usunier, A.~Kirillov, and S.~Zagoruyko, ``End-to-end object detection with transformers,'' in \emph{European conference on computer vision}.\hskip 1em plus 0.5em minus 0.4em\relax Springer, 2020, pp. 213--229.

\bibitem{ren2016faster}
S.~Ren, K.~He, R.~Girshick, and J.~Sun, ``Faster r-cnn: Towards real-time object detection with region proposal networks,'' \emph{IEEE transactions on pattern analysis and machine intelligence}, vol.~39, no.~6, pp. 1137--1149, 2016.

\bibitem{liu2021bev}
M.~Liu and J.~Niu, ``Bev-net: A bird’s eye view object detection network for lidar point cloud,'' in \emph{2021 IEEE/RSJ International Conference on Intelligent Robots and Systems (IROS)}.\hskip 1em plus 0.5em minus 0.4em\relax IEEE, 2021, pp. 5973--5980.

\bibitem{li2024bevnext}
Z.~Li, S.~Lan, J.~M. Alvarez, and Z.~Wu, ``Bevnext: Reviving dense bev frameworks for 3d object detection,'' in \emph{Proceedings of the IEEE/CVF conference on computer vision and pattern recognition}, 2024, pp. 20\,113--20\,123.

\bibitem{peng2023bevsegformer}
L.~Peng, Z.~Chen, Z.~Fu, P.~Liang, and E.~Cheng, ``Bevsegformer: Bird's eye view semantic segmentation from arbitrary camera rigs,'' in \emph{Proceedings of the IEEE/CVF Winter Conference on Applications of Computer Vision}, 2023, pp. 5935--5943.

\bibitem{shi2019pointrcnn}
S.~Shi, X.~Wang, and H.~Li, ``Pointrcnn: 3d object proposal generation and detection from point cloud,'' in \emph{Proceedings of the IEEE/CVF conference on computer vision and pattern recognition}, 2019, pp. 770--779.

\bibitem{qi2019deep}
C.~R. Qi, O.~Litany, K.~He, and L.~J. Guibas, ``Deep hough voting for 3d object detection in point clouds,'' in \emph{proceedings of the IEEE/CVF International Conference on Computer Vision}, 2019, pp. 9277--9286.

\bibitem{9127813}
Y.~Guo, H.~Wang, Q.~Hu, H.~Liu, L.~Liu, and M.~Bennamoun, ``Deep learning for 3d point clouds: A survey,'' \emph{IEEE Transactions on Pattern Analysis and Machine Intelligence}, vol.~43, no.~12, pp. 4338--4364, 2021.

\bibitem{9186684}
H.~Peng, C.~Xue, Y.~Shao, K.~Chen, J.~Xiong, Z.~Xie, and L.~Zhang, ``Semantic segmentation of litchi branches using deeplabv3+ model,'' \emph{IEEE Access}, vol.~8, pp. 164\,546--164\,555, 2020.

\bibitem{WANG2021106373}
C.~Wang, P.~Du, H.~Wu, J.~Li, C.~Zhao, and H.~Zhu, ``A cucumber leaf disease severity classification method based on the fusion of deeplabv3+ and u-net,'' \emph{Computers and Electronics in Agriculture}, vol. 189, p. 106373, 2021.

\bibitem{10.1007/978-3-319-24574-4_28}
O.~Ronneberger, P.~Fischer, and T.~Brox, ``U-net: Convolutional networks for biomedical image segmentation,'' in \emph{Medical Image Computing and Computer-Assisted Intervention -- MICCAI 2015}, N.~Navab, J.~Hornegger, W.~M. Wells, and A.~F. Frangi, Eds.\hskip 1em plus 0.5em minus 0.4em\relax Cham: Springer International Publishing, 2015, pp. 234--241.

\bibitem{Kirillov_2023_ICCV}
A.~Kirillov, E.~Mintun, N.~Ravi, H.~Mao, C.~Rolland, L.~Gustafson, T.~Xiao, S.~Whitehead, A.~C. Berg, W.-Y. Lo, P.~Dollar, and R.~Girshick, ``Segment anything,'' in \emph{Proceedings of the IEEE/CVF International Conference on Computer Vision (ICCV)}, October 2023, pp. 4015--4026.

\bibitem{thomas2019kpconvflexibledeformableconvolution}
H.~Thomas, C.~R. Qi, J.-E. Deschaud, B.~Marcotegui, F.~Goulette, and L.~Guibas, ``Kpconv: Flexible and deformable convolution for point clouds,'' in \emph{2019 IEEE/CVF International Conference on Computer Vision (ICCV)}, 2019, pp. 6410--6419.

\bibitem{hu2020randlanetefficientsemanticsegmentation}
Q.~Hu, B.~Yang, L.~Xie, S.~Rosa, Y.~Guo, Z.~Wang, N.~Trigoni, and A.~Markham, ``Randla-net: Efficient semantic segmentation of large-scale point clouds,'' in \emph{2020 IEEE/CVF Conference on Computer Vision and Pattern Recognition (CVPR)}, 2020, pp. 11\,105--11\,114.

\bibitem{lyu2020learningsegment3dpoint}
Y.~Lyu, X.~Huang, and Z.~Zhang, ``Learning to segment 3d point clouds in 2d image space,'' in \emph{2020 IEEE/CVF Conference on Computer Vision and Pattern Recognition (CVPR)}, 2020, pp. 12\,252--12\,261.

\bibitem{yang20242d3dinterlacedtransformerpoint}
C.-K. Yang, M.-H. Chen, Y.-Y. Chuang, and Y.-Y. Lin, ``{ 2D-3D Interlaced Transformer for Point Cloud Segmentation with Scene-Level Supervision },'' in \emph{2023 IEEE/CVF International Conference on Computer Vision (ICCV)}.\hskip 1em plus 0.5em minus 0.4em\relax Los Alamitos, CA, USA: IEEE Computer Society, Oct. 2023, pp. 977--987.

\bibitem{guo2024samguidedgraphcut3d}
H.~Guo, H.~Zhu, S.~Peng, Y.~Wang, Y.~Shen, R.~Hu, and X.~Zhou, ``Sam-guided graph cut for 3d instance segmentation,'' in \emph{Computer Vision -- ECCV 2024}, A.~Leonardis, E.~Ricci, S.~Roth, O.~Russakovsky, T.~Sattler, and G.~Varol, Eds.\hskip 1em plus 0.5em minus 0.4em\relax Cham: Springer Nature Switzerland, 2025, pp. 234--251.

\bibitem{CDR_v_Caltrans}
``Californians for disability rights, inc.\ v.\ california department of transportation, settlement agreement re class action settlement (case no.\ c 06 5125),'' \url{https://dot.ca.gov/-/media/dot-media/programs/civil-rights/documents/settlement-agreement-a11y.pdf}, accessed: 29 March 2025.

\bibitem{Willits_v_LA}
``Willits v. city of los angeles term sheet as of march 30, 2015,'' \url{https://cao.lacity.gov/sidewalks/Willits_Term%20Sheet_Redacted.pdf}, city Administrative Officer (CAO), Los Angeles. Accessed: 29 March 2025.

\bibitem{lin2014microsoft}
T.-Y. Lin, M.~Maire, S.~Belongie, J.~Hays, P.~Perona, D.~Ramanan, P.~Doll{\'a}r, and C.~L. Zitnick, ``Microsoft coco: Common objects in context,'' in \emph{Computer Vision--ECCV 2014: 13th European Conference, Zurich, Switzerland, September 6-12, 2014, Proceedings, Part V 13}.\hskip 1em plus 0.5em minus 0.4em\relax Springer, 2014, pp. 740--755.

\bibitem{ocsvm}
B.~Schölkopf, J.~C. Platt, J.~Shawe-Taylor, A.~J. Smola, and R.~C. Williamson, ``Estimating the support of a high-dimensional distribution,'' \emph{Neural Computation}, vol.~13, no.~7, pp. 1443--1471, 07 2001.

\bibitem{dbscan}
M.~Ester, H.-P. Kriegel, J.~Sander, and X.~Xu, ``A density-based algorithm for discovering clusters in large spatial databases with noise,'' in \emph{Proceedings of the Second International Conference on Knowledge Discovery and Data Mining}, ser. KDD'96.\hskip 1em plus 0.5em minus 0.4em\relax AAAI Press, 1996, p. 226–231.

\bibitem{iforest}
F.~T. Liu, K.~M. Ting, and Z.-H. Zhou, ``Isolation-based anomaly detection,'' \emph{ACM Trans. Knowl. Discov. Data}, vol.~6, no.~1, Mar. 2012.

\bibitem{fischler1981random}
M.~A. Fischler and R.~C. Bolles, ``Random sample consensus: a paradigm for model fitting with applications to image analysis and automated cartography,'' \emph{Communications of the ACM}, vol.~24, no.~6, pp. 381--395, 1981.

\bibitem{vnet}
F.~Milletari, N.~Navab, and S.-A. Ahmadi, ``V-net: Fully convolutional neural networks for volumetric medical image segmentation,'' in \emph{2016 Fourth International Conference on 3D Vision (3DV)}, 2016, pp. 565--571.

\bibitem{dosovitskiy2021an}
A.~Dosovitskiy, L.~Beyer, A.~Kolesnikov, D.~Weissenborn, X.~Zhai, T.~Unterthiner, M.~Dehghani, M.~Minderer, G.~Heigold, S.~Gelly, J.~Uszkoreit, and N.~Houlsby, ``An image is worth 16x16 words: Transformers for image recognition at scale,'' in \emph{International Conference on Learning Representations}, 2021.

\end{thebibliography}

\end{document}